\documentclass[11pt]{article}

\usepackage[preprint]{acl}

\usepackage{times}
\usepackage{latexsym}
\usepackage{float}  
\usepackage[T1]{fontenc}
\usepackage[utf8]{inputenc}
\usepackage{microtype}
\usepackage{inconsolata}
\usepackage{scalerel}
\usepackage{subcaption}
\usepackage{graphicx}
\usepackage{xcolor}
\usepackage{booktabs}
\usepackage{multirow}
\usepackage{tabularx}
\usepackage{comment}
\usepackage{times}
\usepackage{latexsym}
\usepackage{booktabs}
\usepackage{hyperref} 
\usepackage{tcolorbox}
\tcbuselibrary{skins}
\usepackage{array}
\usepackage{xcolor}
\usepackage{soul}
\usepackage{multicol,multirow}
\usepackage{todonotes}
\usepackage{xspace}
\usepackage[table]{xcolor}
\usepackage{enumitem}
\usepackage{caption}
\usepackage{booktabs}

\renewcommand{\arraystretch}{1.2}

\definecolor{mygold}{HTML}{eeba0a}
\definecolor{mygrey}{HTML}{bac8ca}
\definecolor{mylightgreen}{HTML}{D1F694}
\definecolor{myblue}{HTML}{30C0F0}
\definecolor{myorange}{HTML}{F5AF22}
\definecolor{myred}{HTML}{FF576A}
\definecolor{darkred}{HTML}{91270F}
\definecolor{darkgreen}{HTML}{5E893E}
\definecolor{lightblue}{HTML}{b6ecfc}

\definecolor{chartdarkblue}{HTML}{4C72B0}  
\definecolor{chartlightblue}{HTML}{A0CBE8}  
\definecolor{chartdarkorange}{HTML}{DD8452}
\definecolor{chartlightorange}{HTML}{FFB482}

\definecolor{chartdarkgreen}{HTML}{55A868}  
\definecolor{chartlightgreen}{HTML}{C7E9C0}  

\definecolor{chartdarkred}{HTML}{FF0000}  

\sethlcolor{chartdarkblue}    

\sethlcolor{chartlightblue}    

\sethlcolor{chartdarkorange}  

\sethlcolor{chartlightorange}  

\sethlcolor{chartgreen}   

\sethlcolor{chartred}     

\sethlcolor{mylightgreen}
\usepackage{makecell}
\usepackage{longtable}
\usepackage{amssymb}
\usepackage{pifont}
\definecolor{lightpink}{rgb}{1.0, 0.8, 0.9}

\newcommand{\digitbf}[1]{%
  \colorbox{lightblue}{\textbf{#1}}%
}

\tcbuselibrary{listings, breakable} 

\usepackage{tcolorbox}   
\tcbuselibrary{listings, breakable}  

\usepackage{xcolor}      
\usepackage{listings}    

\usepackage{amssymb}

\newcommand{\plotqacorrect}{\textcolor{chartdarkblue}{$\bullet$}}
\newcommand{\plotqamisleading}{\textcolor{chartdarkblue}{$\circ$}}

\newcommand{\chartqacorrect}{\textcolor{chartdarkorange}{$\blacksquare$}}
\newcommand{\chartqamisleading}{\textcolor{chartdarkorange}{$\square$}}

\newcommand{\chartxcorrect}{\textcolor{chartdarkgreen}{$\blacklozenge$}}
\newcommand{\chartxmisleading}{\textcolor{chartdarkgreen}{$\lozenge$}}

\newcommand{\plotqacorrectdr}{\textcolor{chartdarkblue}{$\blacksquare$}}
\newcommand{\plotqaincorrectdr}{\textcolor{chartlightblue}{$\blacksquare$}}

\newcommand{\chartqacorrectdr}{\textcolor{chartdarkorange}{$\blacksquare$}}
\newcommand{\chartqaincorrectdr}{\textcolor{chartlightorange}{$\blacksquare$}}

\newcommand{\chartxcorrectdr}{\textcolor{chartdarkgreen}{$\blacksquare$}}
\newcommand{\chartxincorrectdr}{\textcolor{chartlightgreen}{$\blacksquare$}}

\newcommand{\zeroshotmarker}{\textcolor{chartdarkblue}{$\blacksquare$}}
\newcommand{\randomshotmarker}{\textcolor{chartdarkorange}{$\blacksquare$}}
\newcommand{\demoshotmarker}{\textcolor{chartdarkgreen}{$\blacksquare$}}

\title{ChartAttack: Testing the Vulnerability of LLMs to Malicious Prompting in Chart Generation}

\author{
  \textbf{Jesus-German Ortiz-Barajas\textsuperscript{\includegraphics[height=2ex]{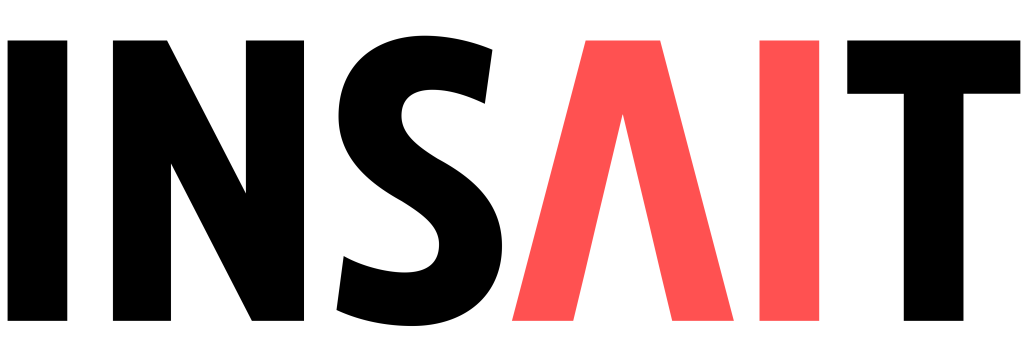}}},
  \textbf{Jonathan Tonglet\textsuperscript{\includegraphics[height=2ex]{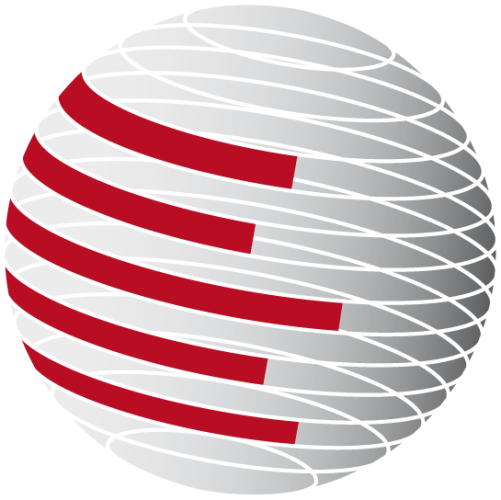},\includegraphics[height=2ex]{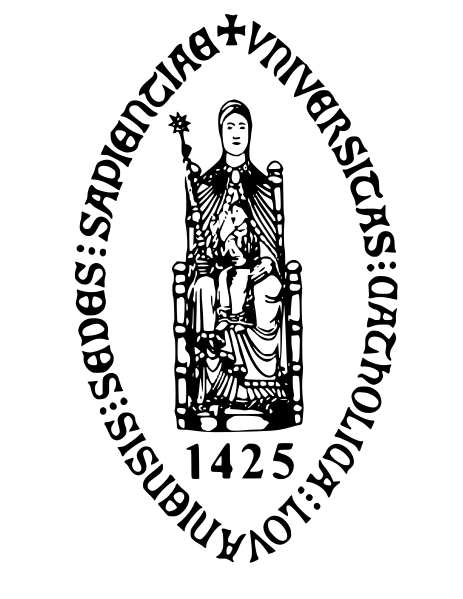}}},
  \textbf{Vivek Gupta\textsuperscript{\includegraphics[height=2ex]{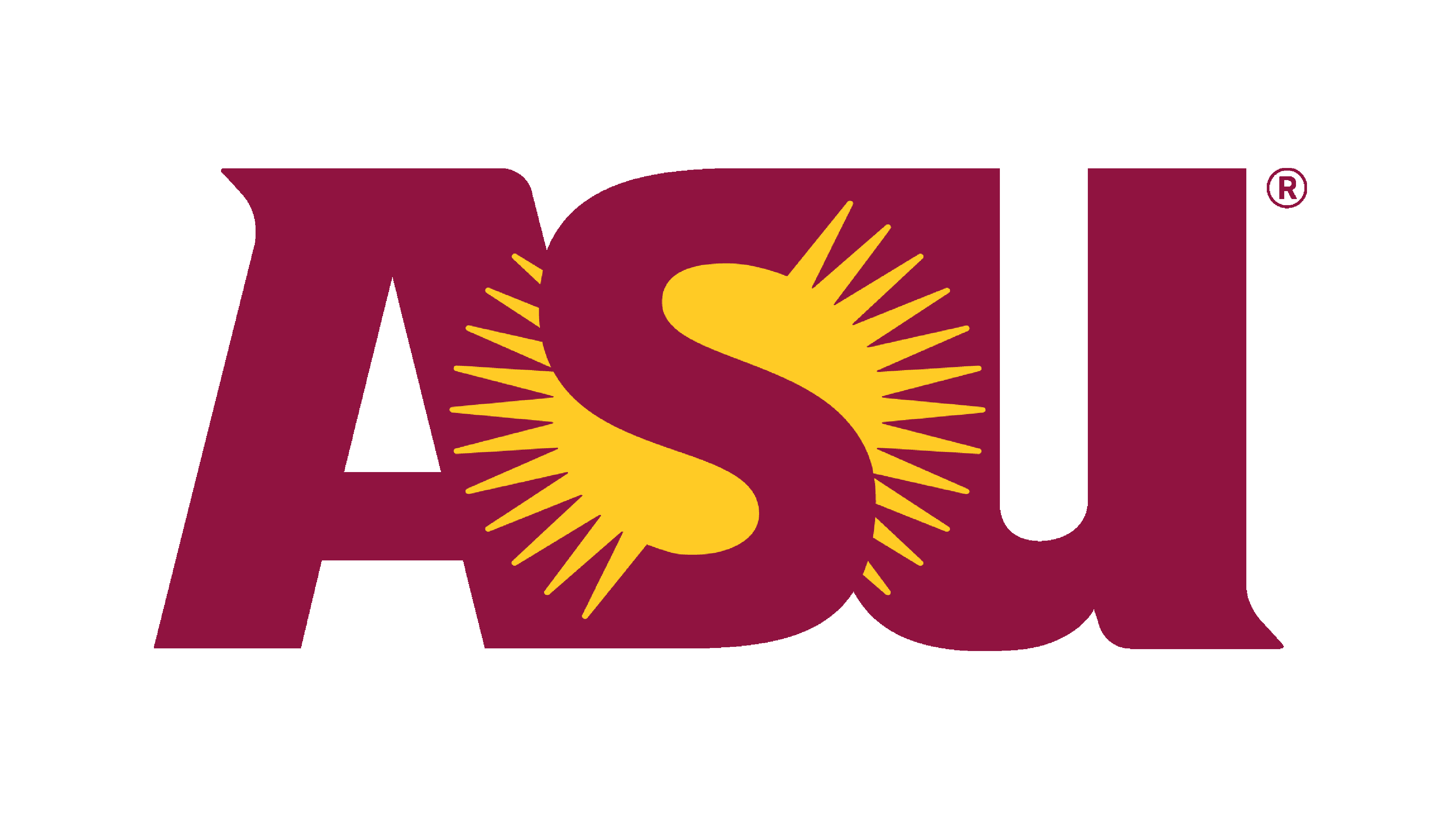}}},
  \textbf{Iryna Gurevych\textsuperscript{\includegraphics[height=2ex]{images/icons/insait_logo.png},\includegraphics[height=2ex]{images/icons/ukp_logo.png}}}
\\
  \textsuperscript{\includegraphics[height=2ex]{images/icons/insait_logo.png}}INSAIT, Sofia University “St. Kliment Ohridski”\\
  \textsuperscript{\includegraphics[height=2ex]{images/icons/ukp_logo.png}}Ubiquitous Knowledge Processing Lab (UKP Lab), Department of Computer Science,\\
TU Darmstadt and National Research Center for Applied Cybersecurity ATHENE\\
\textsuperscript{\includegraphics[height=2ex]{images/icons/ku-leuven.png}}KU Leuven
\\
  \textsuperscript{\includegraphics[height=2ex]{images/icons/asu_logo.png}}Arizona State University
\\
\small{
    \includegraphics[height=2ex]{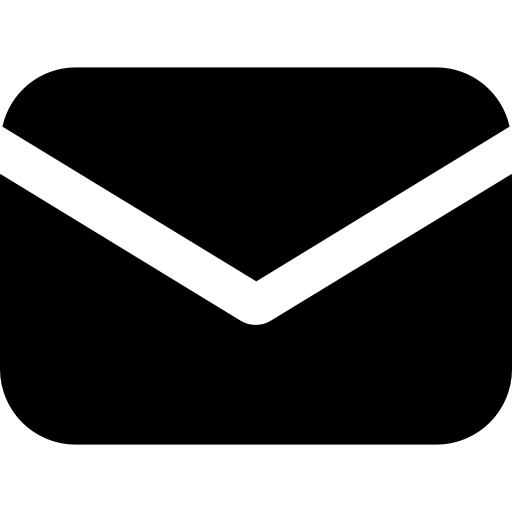} \href{mailto:german.ortiz@insait.ai}{\texttt{german.ortiz@insait.ai}}
}
}

\begin{document}
\maketitle
\begin{abstract}
Multimodal large language models (MLLMs) are increasingly used to automate chart generation from data tables, improving efficiency but introducing new misuse risks. We present ChartAttack, a framework for evaluating how MLLMs use design misleaders to generate charts that induce incorrect interpretations. We also introduce AttackViz, a chart question-answering (QA) dataset labeled with effective misleaders and their induced incorrect answers. ChartAttack reduces MLLM QA accuracy by 17.2 points in-domain and 11.9 points cross-domain. Conditional deception rates show targeted effects: correct answers shift to attacker-intended answers 11.2\% of the time in-domain and 11.7-14.9\% cross-domain, while originally incorrect answers rarely change. A controlled human study shows that ChartAttack-generated charts also reduce human QA performance. Finally, fine-tuning on AttackViz improves in-domain MLLM robustness to misleading charts. Our findings highlight the need for secure, robust MLLM chart generation. Code and data are publicly available on the project website.\footnote{\includegraphics[height=2.0ex]{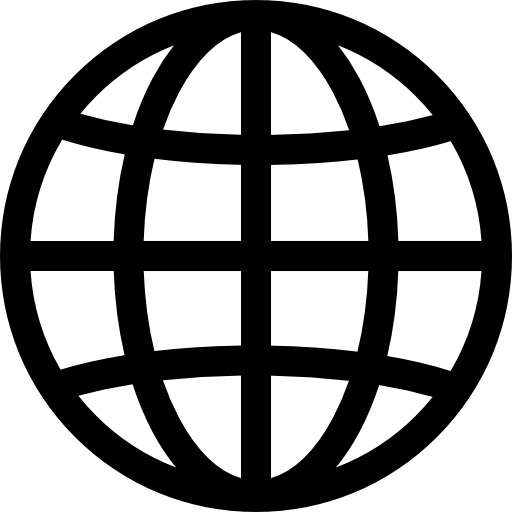} \scriptsize\url{https://chartattack.insait.ai}}
\end{abstract}

\section{Introduction}

\begin{figure}[!ht]
    \centering
    \includegraphics[width=0.80\linewidth]{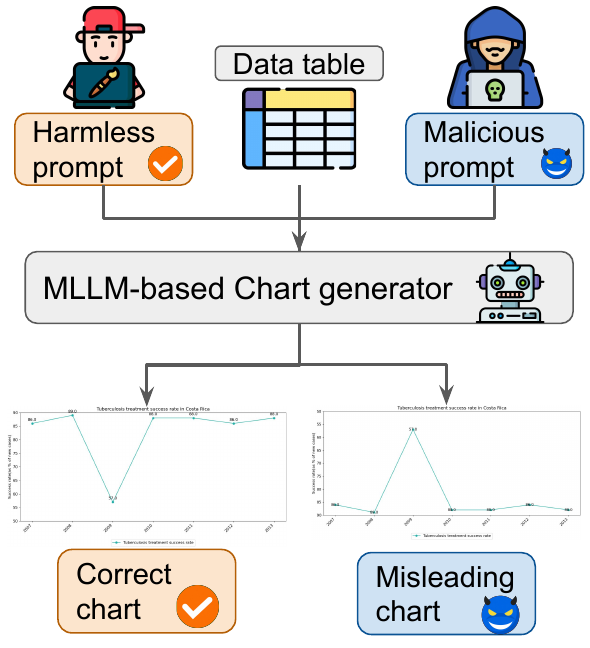}
    \caption{Illustration of dual-use risks in MLLM chart generators: misleading charts that deceive readers.}
    \label{fig:overall_research_problem}
\end{figure}

\begin{figure*}[!ht]
    \centering
    \includegraphics[width=0.80\linewidth]{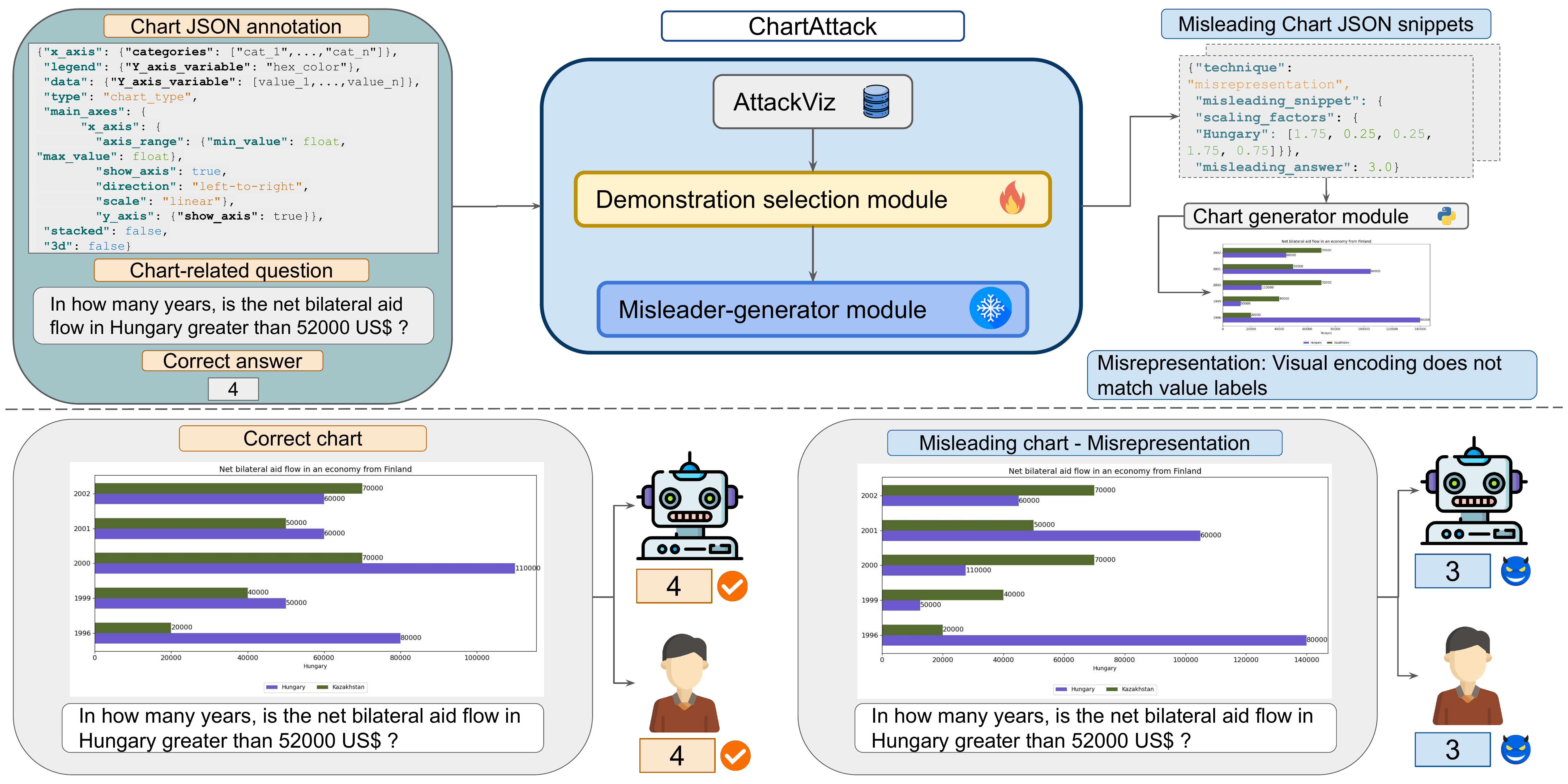}
    \caption{Overview of our ChartAttack framework. The top part shows the generation of misleading charts by the attacker. The bottom part shows the QA evaluation on MLLM and human readers.}
    \label{fig:chartAttack_model}
\end{figure*}

Charts are widely used to communicate complex information across domains such as politics, climate, and healthcare \citep{lauer2020deceptive, huang2025chartUnderstanding}, and play a critical role during crises such as the COVID-19 pandemic \citep{zhang2021Covid, woloshin2023communicating}. However, poorly designed or intentionally manipulated charts can propagate misinformation \citep{huff1993lie, lan2025designflaws}. Misleading charts distort the interpretation of underlying data through misleaders: design choices that violate visualization principles and systematically bias perception or inference, such as inverting axes to reverse perceived trends. Prior work has shown that misleading charts can significantly decrease the QA performance for both human readers \citep{pandey2014persuasive, pandet2015deceptive, obrien2018UsersDeceptive, yang2021truncating, ge2023calvi, rho2024various} and MLLMs \citep{bharti2024chartom,bendeck2024gpt4,chen-etal-2025-unmasking,tonglet2025protecting}.

Chart creation has been democratized by user-friendly tools \citep{pandet2015deceptive}, and designers increasingly use MLLMs for chart generation and analysis \cite{shen2024askhumansaiexploring, ahn2025understandingchatgptoutperformshumans}. While MLLMs simplify legitimate tasks, they can be exploited to generate misleading content at scale \citep{pan-etal-2023-risk,sallami2024deceptiondetectiondualroles,zugecova-etal-2025-evaluation}, including misleading charts (Figure \ref{fig:overall_research_problem}). However, the effectiveness of MLLM-based misleading chart generation and its impact on readers have not been systematically quantified. To address this gap, we investigate three research questions: (1) To what extent can MLLMs generate misleading charts at scale?; (2) How can misleading charts and deceptive intent be systematically annotated?; and (3) How effective are MLLM-generated misleading charts at deceiving humans and MLLMs?

In this work, we present the first systematic study of this jailbreaking attack \citep{wei2023jailbroken, lin-etal-2024-towards-understanding}. We introduce ChartAttack (Figure \ref{fig:chartAttack_model}), a framework that automatically applies misleaders to chart annotations, JSON files specifying the underlying data and chart formatting, to deceive readers with respect to a specific chart question. ChartAttack applies known misleaders that alter chart design without changing the underlying data, enabling the deliberate generation of misleading yet data-consistent charts while keeping the correct answer recoverable. In contrast, modifying values or factual labels would shift the task toward fact-checking or information retrieval, requiring access to the original table or external evidence. We also introduce AttackViz, a multi-label chart QA dataset covering bar (horizontal and vertical) and line charts. Each instance contains chart annotations, an associated question, and a set of misleaders with annotations specifying how each is applied and the incorrect answers it causes.

We evaluate ChartAttack on both MLLMs and human readers. It reduces average MLLM QA accuracy by 17.2 percentage points (pp) in-domain and 11.9 pp cross-domain. In a controlled human study, misleading charts generated by ChartAttack significantly reduced chart QA accuracy. We further show that AttackViz can be used to fine-tune an MLLM for improved in-domain robustness to misleading charts, increasing test-set performance by 10.51 pp. 

We summarize our contributions as follows: (1) We introduce ChartAttack, the first framework for automatically generating misleading charts through reproducible and parameterizable misleaders designed to induce targeted misinterpretations. (2) We present AttackViz, a chart QA dataset with structured annotations containing both original chart annotations with correct answers and misleading chart annotations with their resulting incorrect answers. (3) We extensively evaluate misleading chart attacks on MLLMs and conduct a human study showing that MLLM-generated misleading charts can deceive both models and human readers.

\section{Related work}
\paragraph{Misleading charts and MLLMs.}
Prior work has focused on two directions. The first one investigates MLLMs' ability to interpret charts and their vulnerability to misleading designs in a QA setting \cite{bharti2024chartom, bendeck2024gpt4, chen-etal-2025-unmasking, zeng2025AdvancingMLLMsinChartQA, tonglet2025protecting, mahbub2025perilschartdeceptionmisleading, pandey2025Becnhmarking}. Some works proposed inference-time strategies to reduce QA errors, with moderate success \citep{tonglet2025protecting,chen-etal-2025-unmasking}. The second one leverages MLLMs to detect and correct misleading charts \cite{alexander2024GPT4, lo2025LLMsatDetecting, kim2025automatedPipelineMisleading, gangwar2025automatedvisualizationmakeoversllms, das2025MisvizFix, tonglet2025chartlyingmeautomating}. By contrast, our work analyzes whether MLLMs can be misused to generate misleading charts that can effectively deceive humans and other MLLMs.

\paragraph{Jailbreak attacks.}
The widespread use of MLLMs has intensified concerns about jailbreaking, where adversarial prompts bypass safety mechanisms to induce harmful or misleading outputs \citep{lin-etal-2024-towards-understanding}. A common class of attacks relies on template completion, exploiting MLLMs’ role-playing and contextual reasoning abilities. Within this class, scenario nesting attacks craft deceptive contexts that gradually steer models toward unsafe behaviors \citep{ding-etal-2024-wolf, yuan2024gpt, cui-etal-2025-exploring}. Another template completion approach is context-based attacks, where adversarial examples are embedded directly into the prompt context to exploit in-context learning and override safety constraints \citep{li-etal-2023-multi-step, globerson2024Many-shot, zheng2024improved, pernisi-etal-2024-compromesso}. We present the first systematic study of a scenario-nesting-based malicious prompting attack for generating misleading charts and evaluating its effectiveness on humans and MLLMs. ChartAttack combines adversarial in-context demonstrations with scenario nesting, an established jailbreak-prompt technique \cite{ding-etal-2024-wolf}, while evaluating malicious instruction following rather than requiring demonstrated refusal bypass \cite{xia-etal-2025-threat}.

\section{ChartAttack framework}
\label{sec:chartAttack}
ChartAttack (Figure~\ref{fig:chartAttack_model}) is a framework for generating misleading charts by applying misleaders to chart annotations to induce incorrect answers to chart-related questions. The input consists of chart annotations with data and formatting specifications, along with a question and its correct answer.
The framework has two components. The Demonstration Selection module retrieves similar examples for few-shot prompting. The Misleading Generator module takes the chart annotations, question, correct answer, and retrieved demonstrations, and outputs a list of misleaders, modified annotation snippets, and misleading answers. Each misleading answer is plausible but incorrect while preserving the correct answer’s type and units. This enables evaluation of whether a misleader induces a targeted misinterpretation rather than random errors.

\paragraph{Demonstration selection module.}
The effectiveness of in-context learning depends on the quality of selected examples \citep{liu-etal-2022-makes,wang-etal-2024-learning}. To retrieve relevant demonstrations from a large corpus, we fine-tune an SBERT model \cite{reimers-2019-sentence-bert} using Multiple Negative Ranking Loss \cite{henderson2017efficient}. A demonstration–input pair is considered positive if their sets of misleaders match exactly.
For both corpus candidates and input instances, SBERT encodes the concatenation of the question and its chart JSON annotation. Top-$k$ demonstrations are retrieved using cosine similarity and included in the prompt of the Misleading Generator module to guide misleading chart generation.
We train a separate retriever for each chart type because each type is affected by a different set of misleaders and has distinct chart semantics. Experimental results are reported in §~\ref{subsubsec:demonstration_ablation}, and the dataset creation process for this module is detailed in Appendix \ref{sec:appendix_demonstration_selection_training_dataset}.

\paragraph{Misleader-generator module.}
The module applies misleaders to chart JSON annotations to induce incorrect answers to associated questions. We use code-based instruction-tuned MLLMs to modify the annotations, as they outperform general MLLMs on structured reasoning tasks \citep{madaan-etal-2022-language}. We select models based on performance on the Human-Eval benchmark \cite{chen2021HumanEval} and use a few-shot prompting strategy with $k$ demonstrations. The module takes three inputs: (1) chart annotations containing the data and basic formatting specifications, (2) the associated question, and (3) similar examples retrieved by the Demonstration Selection module.
We conduct ablation studies on the demonstration selection and misleader-generator modules to identify the optimal loss function, downsampling strategy, code-based instruction MLLM, retrieval strategy, and number of demonstrations. Detailed results are reported in Appendix \ref{sec:appendix_ablations}.

We use a shared prompt template across chart types. In a single inference call, the model performs three structured steps: (i) select misleaders compatible with the chart and context; (ii) specify minimal modifications to apply each misleader without altering other chart elements; and (iii) produce a misleading answer based on the applied misleader. This design ensures consistent generation of misleading chart variants. Details are provided in Appendix \ref{sec:appendix_misleader-generator_prompt}.

\section{AttackViz corpus}

\begin{figure*}[!ht]
    \centering
    \includegraphics[width=0.925\linewidth]{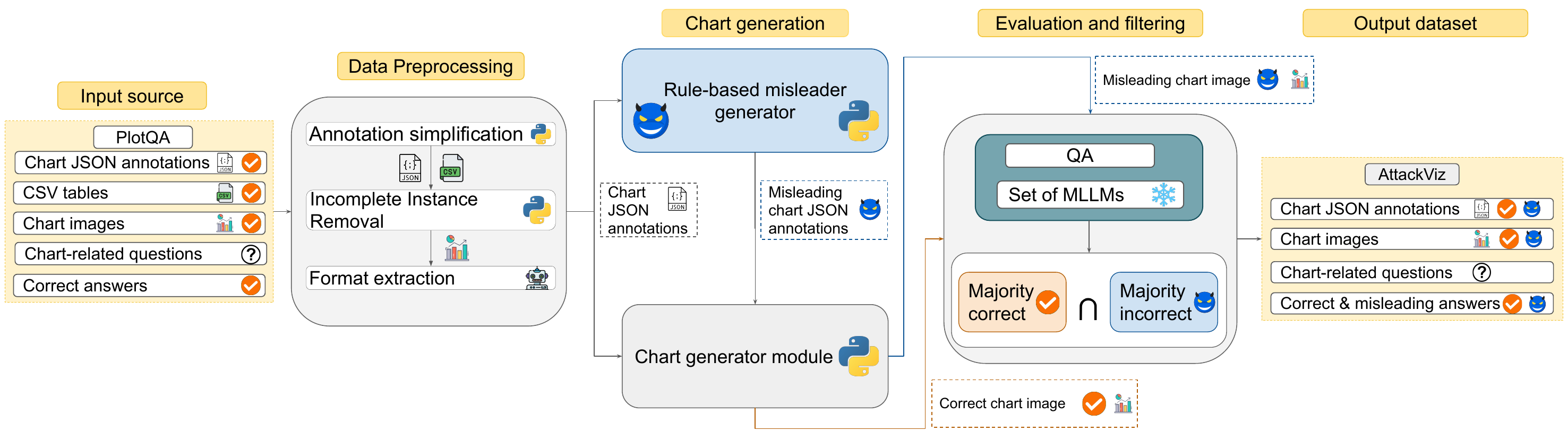}
    \caption{Pipeline to create the AttackViz corpus.}
    \label{fig:attackViz}
\end{figure*}

We create the AttackViz corpus to support ChartAttack. It serves two main purposes: (i) as a candidate pool for the Demonstration selection module, and (ii) to evaluate how effectively our model can deceive MLLMs or humans in a chart QA setting. Figure \ref{fig:attackViz} illustrates the corpus creation pipeline.

\paragraph{Input source.}
We construct AttackViz using PlotQA \cite{methani2020plotqa}. This dataset provides train, validation, and test splits. Each instance contains chart images, JSON annotation files with the underlying data and metadata (e.g., title, axis labels, and chart type), a CSV file representing the data table, and associated question-answer pairs. The plots are generated from real-world online sources such as World Bank Open Data, Open Government Data, and the Global Terrorism Database.

\paragraph{Data preprocessing.}
First, we simplify the chart JSON annotations to reduce complexity and improve readability for chart generation and misleader selection. We remove bounding boxes, label coordinates, and figure geometry, and reorganize the remaining content into lists of categories, values, legends, and colors. We then verify consistency with the CSV data tables to ensure charts accurately reflect the underlying data.
Finally, we use Phi-3.5-vision \cite{abdin2024phi3technicalreporthighly}, a lightweight MLLM, to extract chart format information: we determine whether charts contain grids, bands, and whether horizontal or vertical bar charts are stacked. This produces a simplified, data-consistent, and format-rich JSON annotation for each chart. We randomly subsample 400 images per chart type for each partition (train, validation, test) and retain five questions per chart to cover all PlotQA question types. 

\paragraph{Rule-based misleading chart generation and chart coverage.}
We generate misleading charts using a rule-based system implementing 11 misleaders from the taxonomy of \citet{lo2022misinformed}. Table \ref{tab:misleading_definitions} summarizes the selected techniques. Misleaders are chosen according to six criteria: (1) at least five occurrences in real-world examples; (2) previously studied in misleading chart QA \citep{ge2023calvi,bharti2024chartom} or visualization design-support research \citep{lo2023change}; (3) the correct answer to the associated question remains unchanged after applying the misleader; (4) the technique violates grammar rules; (5) the underlying data table remains correct; and (6) the misleader can be implemented in Python. Criteria (3)–(5) ensure misleaders use controlled visualization-level perturbations rather than data modification, isolating the effect of a misleading design while keeping the correct answer recoverable from the underlying data. Appendix \ref{sec:appendix_attackViz_misleader} reports which criteria are satisfied by each taxonomy misleader. 

We focus on bar (horizontal and vertical) and line charts, which account for 64\% of the taxonomy and 49\% of real-world misleading visualizations in the MisViz benchmark \cite{tonglet2025chartlyingmeautomating}. While ChartQA and ChartX include additional chart types, these are affected by fewer misleaders. For example, pie charts are affected by two and map charts by three of the eleven techniques. Consequently, these chart types are less relevant due to their smaller misleading-design search space. Our selection aligns with studies identifying truncated axes, 3D effects, dual axes, and visual misrepresentation as frequent real-world issues, while pie-chart misuse, map distortion, and misleading titles are less prominent \cite{lo2022misinformed, lo2025LLMsatDetecting}.

We implement the system in Python using Matplotlib \citep{hunter2007Matplotlib}. It applies misleaders by modifying chart JSON annotations and rendering the resulting charts, while unmodified annotations produce the correct versions. Operating at the annotation level also enables compatibility with other visualization libraries.

\begin{table*}[t]
\centering
\tiny
\rowcolors{2}{gray!10}{white}
\begin{tabularx}{\textwidth}{l m{0.6\textwidth} >{\centering\arraybackslash}X}
\rowcolor{yellow!45}
\textbf{Misleader} & \textbf{Definition} & \textbf{Affected chart types} \\
\midrule

Dual axis & Two independent axes are layered with inappropriate scaling, creating a misleading narrative about the relationship between them. & 
\includegraphics[height=1.5em]{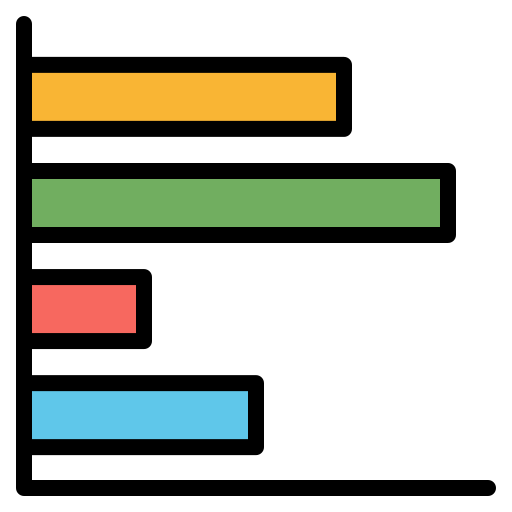} \quad
\includegraphics[height=1.5em]{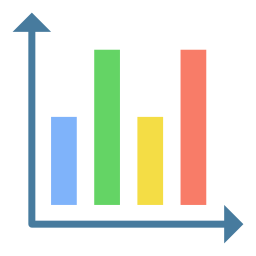} \quad
\includegraphics[height=1.5em]{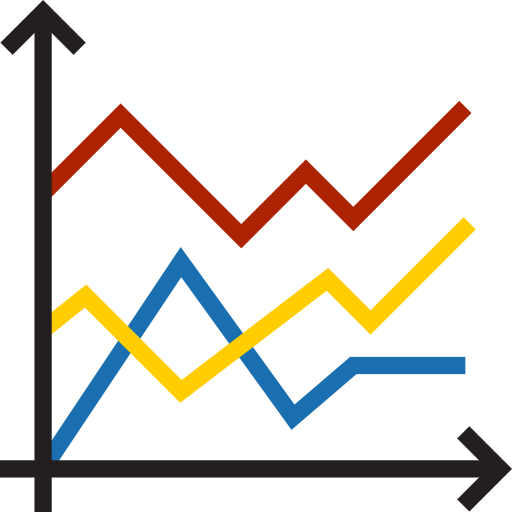} \\

Inverted axis & An axis oriented in an unconventional direction, reversing the perception of the data and potentially confusing the audience. &
\includegraphics[height=1.5em]{images/icons/h_bar_icon.png} \quad
\includegraphics[height=1.5em]{images/icons/v_bar_icon.png} \quad
\includegraphics[height=1.5em]{images/icons/line_icon.png} \\

Inappropriate use of log scale & A logarithmic scale applied to non-exponential data, leading to misinterpretation. &
\includegraphics[height=1.5em]{images/icons/h_bar_icon.png} \quad
\includegraphics[height=1.5em]{images/icons/v_bar_icon.png} \quad
\includegraphics[height=1.5em]{images/icons/line_icon.png} \\

Inappropriate axis range & The axis range is either too broad or too
narrow to accurately visualize the data, allowing changes to be minimized or maximized depending on
the author’s intention. &
\includegraphics[height=1.5em]{images/icons/h_bar_icon.png} \quad
\includegraphics[height=1.5em]{images/icons/v_bar_icon.png} \quad
\includegraphics[height=1.5em]{images/icons/line_icon.png} \\

Inappropriate item order & The items are arranged in an unconventional order, misleading the
audience or creating confusion. &
\includegraphics[height=1.5em]{images/icons/h_bar_icon.png} \quad
\includegraphics[height=1.5em]{images/icons/v_bar_icon.png} \quad
\includegraphics[height=1.5em]{images/icons/line_icon.png} \\

Misrepresentation & Visual encoding does not match value labels, e.g., values drawn disproportionately or not to scale, intentionally or unintentionally misrepresenting the data. &
\includegraphics[height=1.5em]{images/icons/h_bar_icon.png} \quad
\includegraphics[height=1.5em]{images/icons/v_bar_icon.png} \quad
\includegraphics[height=1.5em]{images/icons/line_icon.png} \\

Inappropriate use of stacked & Too many layers are stacked, making the visualization difficult to interpret. &
\includegraphics[height=1.5em]{images/icons/h_bar_icon.png} \quad
\includegraphics[height=1.5em]{images/icons/v_bar_icon.png} \\

3D & Objects closer in perspective appear larger despite being the same size in 3D, causing misleading perception. &
\includegraphics[height=1.5em]{images/icons/h_bar_icon.png} \quad
\includegraphics[height=1.5em]{images/icons/v_bar_icon.png} \\

Ineffective color scheme & A color scheme that does not effectively represent data, such as rainbow colors for sequential data or categorical colors for continuous data. &
\includegraphics[height=1.5em]{images/icons/h_bar_icon.png} \quad
\includegraphics[height=1.5em]{images/icons/v_bar_icon.png} \\

Truncated axis & The axis does not start from zero or is truncated in the middle,
resulting in an exaggerated difference between the two bars.&
\includegraphics[height=1.5em]{images/icons/h_bar_icon.png} \quad
\includegraphics[height=1.5em]{images/icons/v_bar_icon.png} \\

Inappropriate use of line & A line chart used in an unconventional way or in a way that misrepresents data, e.g., encoding a categorical variable on an axis or placing time on the y-axis. &
\includegraphics[height=1.5em]{images/icons/v_bar_icon.png} \\

\bottomrule
\end{tabularx}
\caption{Definitions of the misleaders used to build the AttackViz corpus \citep{lo2022misinformed}.}
\label{tab:misleading_definitions}
\end{table*}

\paragraph{Evaluation and filtering process.}
\label{sec:filtering}
We perform chart QA on each correct chart and its misleading counterpart generated by our rule-based system and evaluate performance using relaxed accuracy \cite{masry2022chartqa,methani2020plotqa}. We use three instruction-tuned MLLMs selected based on ChartQA test-set performance \citep{masry2022chartqa}: QwenVL 2.5-32B \cite{bai2025qwen2}, InternVL 3.0-38B \cite{zhu2025internvl3}, and KimiVL-A3B \cite{team2025kimi}. We retain instances where the majority of models answer correctly on the original chart but incorrectly on the misleading chart. A consistency filter ensures that errors are attributable to the misleader: numeric answers must have a standard deviation below 0.5, while textual answers must share a majority identical incorrect response. The final misleading answer is obtained by averaging incorrect numeric responses or taking the majority vote for textual responses. A detailed analysis of the filtering process is provided in Appendix \ref{sec:appendix_filtering_and_selection_effects}.

\paragraph{Cross-domain extension.}
We apply the same pipeline to two additional datasets to extend our corpus to new domains. ChartQA \citep{masry2022chartqa} contains real-world charts from Statista, Pew Research Center, Our World in Data, and the OECD; due to missing or inconsistent annotations, we merge all instances into a single test set (Appendix~\ref{sec:appendix_attackViz_cross}). ChartX \citep{xia2025chartx} provides chart types that can be directly converted into structured data and spans commerce, industry, society, culture, and lifestyle.

The resulting AttackViz corpus is multi-label. Each instance contains a simplified, data-consistent, and format-rich JSON annotation, a question, and a list of misleaders, each specifying its application in JSON and the corresponding misleading answer. Dataset statistics and examples of all chart types and misleaders appear in Appendix \ref{sec:appendix_attackViz_statistics}, with a detailed comparison to prior work on misleading chart understanding in Appendix \ref{subsec:appendix_positioning}.

\section{Experiments}

\begin{figure*}[!ht]
\centering
\includegraphics[width=\linewidth]{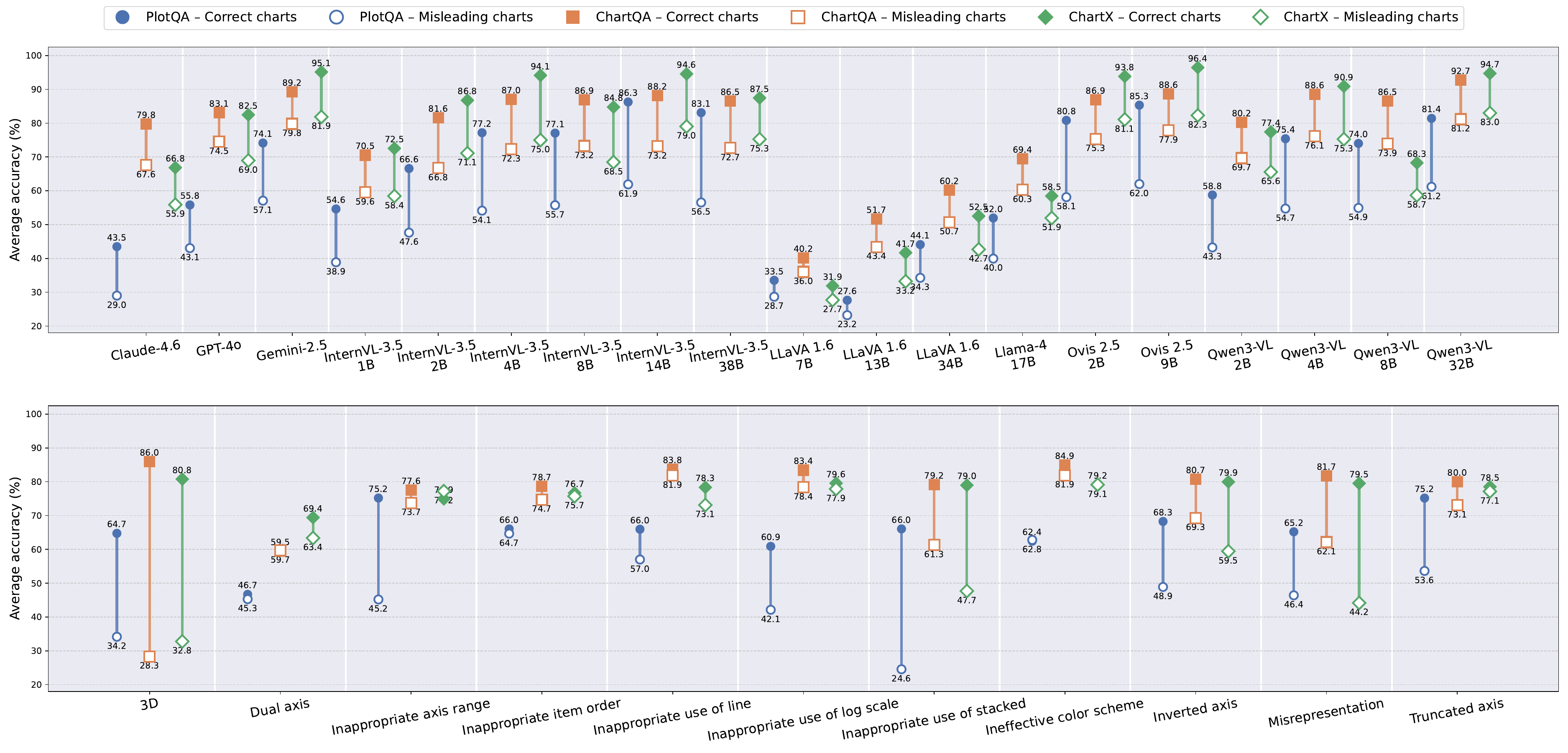}


\caption{
Average accuracy on AttackViz.
\textbf{Top:} Results by model.
\textbf{Bottom:} Results by misleader.
Datasets are encoded by marker shape and color:
\textbf{PlotQA} uses blue circles (\plotqacorrect, \plotqamisleading),
\textbf{ChartQA} uses orange squares (\chartqacorrect, \chartqamisleading), and
\textbf{ChartX} uses green diamonds (\chartxcorrect, \chartxmisleading).
Filled markers (\plotqacorrect, \chartqacorrect, \chartxcorrect) indicate accuracy on correct charts, while hollow markers (\plotqamisleading, \chartqamisleading, \chartxmisleading) indicate accuracy on misleading charts.
}

\label{fig:victim_performance_accuracy}
\end{figure*}

\subsection{Experimental setup}\label{subsection:experimental_setup}

\paragraph{Dataset.} We evaluate ChartAttack using the clean chart specifications, questions, and correct answers from the AttackViz test splits derived from PlotQA, ChartQA, and ChartX. For each test instance, ChartAttack independently generates the applicable misleader, modified chart specification, and misleading answer; the rule-based misleader annotations and misleading variants associated with the test instance are not provided to the framework. We evaluate both in-domain and cross-domain generalization. In the in-domain setting, demonstrations are retrieved from the PlotQA training split, and evaluation is performed on the PlotQA test split. In the cross-domain setting, demonstrations are again retrieved from the PlotQA training split, while evaluation is performed on the ChartQA and ChartX test splits.

\paragraph{Models.} Following prior work on MLLM vulnerabilities to misleading charts \cite{tonglet2025protecting}, we evaluate 16 open-weight instruction-tuned models: Ovis-2.5 (2B, 9B) \cite{lu2025ovis2}, InternVL-3.5 (1B, 2B, 4B, 8B, 14B, 38B) \cite{wang2025internvl3}, LLaVA-1.6 (7B, 13B, 34B)  \cite{liu2024improved}, Qwen3-VL (2B, 4B, 8B, 32B) \cite{yang2025qwen3}, and LLaMA-4 (17B-16E) \cite{meta2025llama4}. Open-weight models are loaded using HuggingFace Transformers \cite{wolf2019huggingface}. We also evaluate three closed models: GPT-4o \cite{alexander2024GPT4}, Gemini-2.5-Flash \cite{comanici2025gemini25pushingfrontier}, and Claude-4.6-Sonnet \cite{anthropic2026claude46} accessed through OpenRouter.\footnote{https://openrouter.ai/} We exclude chart-specialized MLLMs, as recent general-purpose MLLMs outperform them \cite{NGUYEN2026104608,wang2025internvl3} \footnote{All evaluated models are newer than those used to construct the AttackViz corpus in Section 4.}.

\paragraph{Evaluation metrics.} We evaluate each model under two settings: (i) with the correct chart and (ii) with the misleading chart generated by ChartAttack. Following prior work \cite{methani2020plotqa, masry2022chartqa}, we report relaxed accuracy as the primary metric. It is equivalent to accuracy, except that numerical answers are considered correct if they fall within a 5\% tolerance of the ground-truth value. We also introduce two deception-rate metrics. We compute them analogously to relaxed accuracy, but using the misleading answer produced by ChartAttack as the target answer instead of the correct answer. Deception rate (originally correct) measures the percentage of instances where a model answers correctly on the correct chart but outputs the misleading answer on the misleading chart. Deception rate (originally incorrect) measures the percentage of instances where an incorrect answer is replaced by the misleading answer, indicating whether misleading charts reinforce existing errors. Achieving high deception rates is challenging because predictions must match the targeted misleading answer under relaxed-accuracy criteria, evaluating whether the attack induces the intended misinterpretation.

\subsection{MLLM-based evaluation results}\label{subsec:mllm_vulnerabilities}
\begin{figure*}[!ht]
    \centering
    \includegraphics[width=\linewidth]{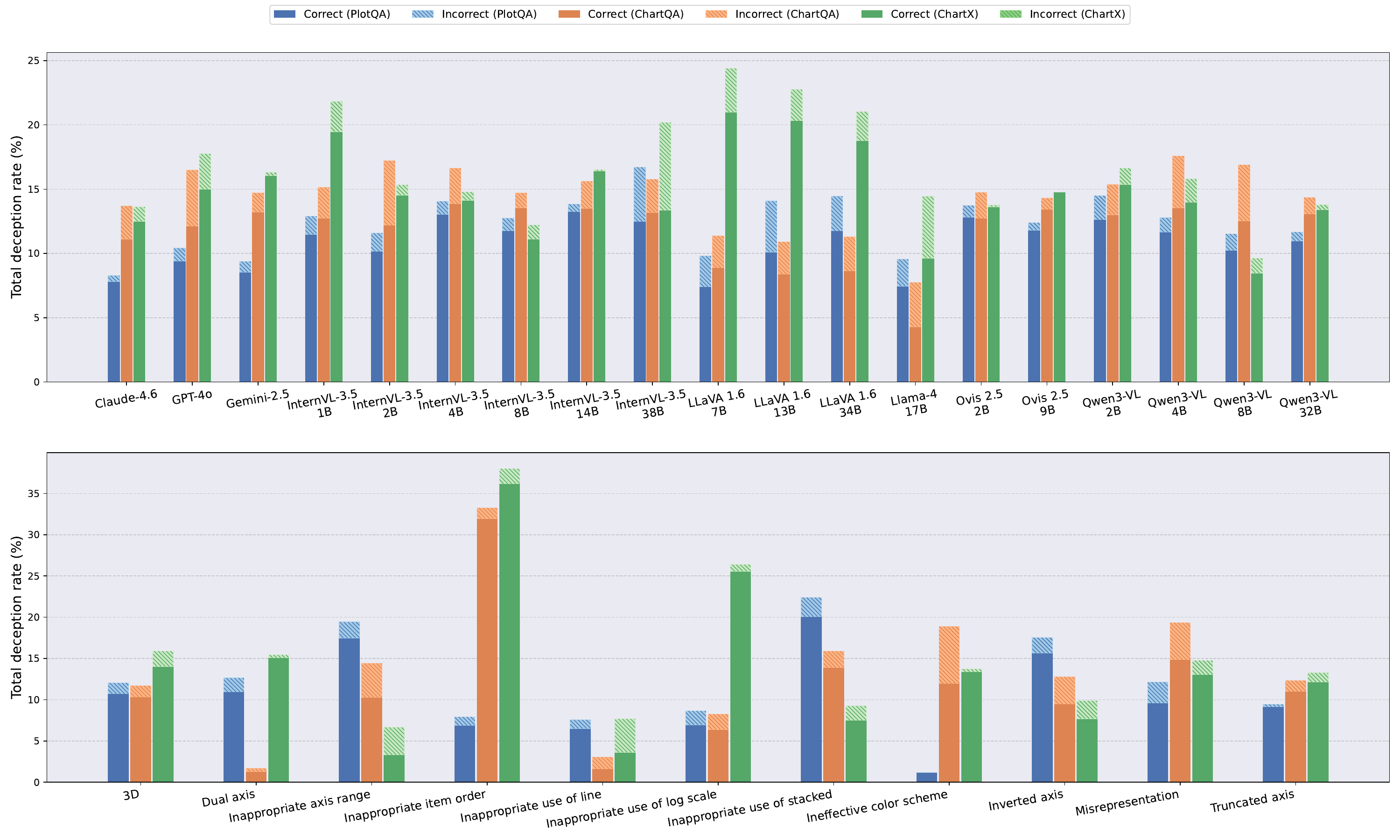}


\caption{
Average deception rate (DR) on AttackViz.
\textbf{Top:} Results by model.
\textbf{Bottom:} Results by misleader.
Datasets are encoded by color:
\textbf{PlotQA} uses blue (\plotqacorrectdr, \plotqaincorrectdr),
\textbf{ChartQA} uses orange (\chartqacorrectdr, \chartqaincorrectdr), and
\textbf{ChartX} uses green (\chartxcorrectdr, \chartxincorrectdr).
Solid bar segments indicate DR on correct, while hatched bar segments indicate DR on incorrect.
Bars are stacked to represent conditional deception rates.
}

\label{fig:victim_performance_deception_rate}
\end{figure*}

We first evaluate the effectiveness of ChartAttack in degrading chart question-answering performance of MLLMs under two settings: in-domain and cross-domain. Figure \ref{fig:victim_performance_accuracy} reports average accuracy, with the top panel showing results for correct and misleading charts for the 19 evaluated models, ordered by parameter size, and the bottom panel aggregating the same metrics by misleader; Figure \ref{fig:victim_performance_deception_rate} reports average conditional deception rates for misleading charts, with the top panel showing results for the 19 models, ordered by parameter size, and the bottom panel aggregating results by misleader. These results reveal the following findings.

\paragraph{In-domain findings.} 
All models perform worse on misleading charts, with accuracy drops ranging from 4.4 to 26.6 pp (17.2 pp on average). Larger drops generally occur among stronger models: LLaVA-1.6 variants (28–44\% accuracy on correct charts) decline by 4–10 pp. Mid-range models, including Claude-4.6, GPT-4o, Gemini-2.5, Qwen3-VL, and smaller InternVL variants (52–77\%), drop by 12–21 pp. Higher-performing models such as InternVL-3.5 (14B/38B) and Ovis-2.5 (2B/9B) achieve 80–86\% accuracy but exhibit larger declines of 22–27 pp. Conditional deception rates show that most errors arise when correct answers shift to attacker-generated misleading answers (11.2\% on average), whereas originally incorrect answers rarely change (1.7\%).

The impact varies across model families. LLaVA-1.6 drops increase from 4.9 pp (7B) to 9.8 pp (34B) despite the size difference. InternVL-3.5 shows stronger scaling effects, with drops rising from 15.8 pp (1B) to 21.3 pp (8B) and 26.6 pp (38B). Although the overall accuracy drop tends to increase with model size, this trend is not strictly monotonic for two reasons. First, the absolute drop depends on baseline performance: larger checkpoints solve more clean-chart instances and therefore have more correct predictions that can be lost, even though their conditional failure rate on initially correct instances is generally lower. Second, checkpoints differ in their sensitivity to specific misleading visualizations. For example, the 8B model exhibits smaller accuracy drops than the 4B model for dual axes (4.2 vs. 12.5 pp) and truncated axes (3.4 vs. 10.1 pp), but a larger drop for inappropriate stacking (53.1 vs. 47.3 pp). These results indicate that parameter count alone does not determine robustness to misleading charts. Other strong multimodal models including GPT-4o, Gemini-2.5, Claude-4.6, and Qwen3-VL show substantial but intermediate drops.

At the misleader level, perceptual manipulations produce the strongest effects. Stacked charts, 3D charts, and inappropriate log scales reduce accuracy to 24.6\%, 34.2\%, and 42.1\%, corresponding to drops of 41.5 pp, 30.6 pp, and 18.8 pp, and higher deception rates on originally correct answers (20.0\%, 10.7\%, and 6.9\%). Misrepresentation and inverted axes cause moderate declines (18.8 pp and 19.4 pp) with deception rates of 9.6\% and 15.6\%. Inappropriate line charts have a smaller impact (9.0 pp drop, 6.5\% deception), while ineffective color schemes have minimal effect (0.4 pp increase, 1.2\% deception). Dual axes lead to only a 1.4 pp overall decline; however, when applied effectively, they can still produce misleading answers generated by ChartAttack (10.9\% deception). We include a preliminary compositional-misleader experiment in Appendix \ref{sec:appendix_compositional_misleaders}, limited to AttackViz-PlotQA vertical bar charts, one generator model, one victim model, and three predefined misleader combinations.
\paragraph{Cross-domain findings.}
Consistent with in-domain results, accuracy on misleading charts drops in ChartQA and ChartX by 4.2–19.1 pp across models, with average declines of 11.5 pp and 12.3 pp, respectively. High-performing models are not immune: InternVL-3.5 (14B/38B), Ovis-2.5 9B, GPT-4o, Gemini-2.5, and Claude-4.6 also show substantial degradation. Conditional deception rates remain low: 11.7\%/14.9\% for originally correct answers and 2.7\%/1.9\% for originally incorrect answers on ChartQA/ChartX.

At the technique level, several trends from the in-domain experiments persist. 3D remains the most impactful technique, reducing accuracy to 27.9\% and 22.7\% (drops of 55.1 pp and 61.4 pp) on ChartQA and ChartX, with deception rates on originally correct answers of 4.2\% and 5.0\%. Misrepresentation follows, with 56.9\% and 54.1\% accuracy (-20.6 pp and -23.0 pp) and deception rates of 6.1\% and 6.3\%. Inappropriate stacked bars also remain effective, yielding 59.3\% and 55.9\% accuracy (-17.2 pp and -20.5 pp) and deception rates of 3.7\% and 4.3\%. In contrast, Dual axis and Ineffective color scheme remain largely ineffective, producing negligible changes (+0.2 pp and 0.0 pp) and small drops (1.8 pp and 2.6 pp). Inappropriate line charts, log scales, axis ranges, and truncated axes show a different pattern: although effective in PlotQA, they cause only modest drops in ChartQA and ChartX, suggesting their impact depends more strongly on dataset characteristics. These results indicate that some misleaders generalize across domains, while others are more sensitive to chart and question semantics. We analyze the cross-domain performance in Appendix \ref{sec:appendix_cross_domain_results} and the performance drops across families, chart types, and misleaders in both settings in Appendix \ref{sec:appendix_mllm_analysis}.

\paragraph{Random baseline comparison.}
We compare ChartAttack with a random baseline. For each chart, the baseline samples the same number of applicable techniques as ChartAttack and applies the same construction rules. Both use the AttackViz-PlotQA test split. On InternVL-3.5-14B, ChartAttack reduces relaxed accuracy by 25.3 pp versus 22.7 pp for random selection and causes 13.2\% targeted deception, converting correct answers to the attacker-specified answer. The random baseline has no target answer, so targeted deception is undefined. ChartAttack's advantage is concentrated in techniques requiring parameter selection, especially truncated axis and inappropriate axis range; single-flag techniques such as 3D, log scale, and inverted axis often produce identical charts under both methods. Its contribution is learned technique and parameter selection, plus targeted deception, rather than stronger application of each misleader.

\subsection{Human evaluation}
We conduct a two-phase human study with 48 participants in a controlled setting to isolate the effect of misleading chart design on chart QA performance, consistent with established practice in human evaluation \cite{mehandru-etal-2023-physician, zhang-etal-2023-biasx, briakou-etal-2023-explaining, kantharaj-etal-2022-chart}. To avoid priming effects from exposure to both correct and misleading charts, we use a between-subjects design, following prior work on deception and AI-generated misinformation \cite{tonglet-etal-2025-cove,wang-etal-2025-llms-reopened, wongkamjan-etal-2025-trust}. Both groups view correct charts in the first phase, while only the experimental group views misleading charts in the second phase. We analyze responses at the individual question level using logistic regression with participant-clustered robust standard errors \cite{schuffVAV23}. Participants exposed to misleading charts had significantly lower odds of answering correctly than participants exposed to correct charts (OR = 0.266, 95\% CI [0.197, 0.357], p < 0.001), corresponding to approximately 73\% lower odds of a correct response. Baseline performance was comparable between groups, while second-phase accuracy decreased from 88.3\% in the control group to 71.9\% in the experimental group. Additional details are provided in Appendix \ref{sec:appendix_human_eval}.

\section{Mitigation strategies}
\begin{table}[t]
\centering

\rowcolors{2}{gray!10}{white}
\resizebox{\columnwidth}{!}{%
\begin{tabular}{lcc}
\rowcolor{yellow!45}
\textbf{Attacker Setting} & \textbf{No Guard ASR\_eff} & \textbf{Guard ASR\_eff} \\
\midrule

Qwen (line, Zero-shot) & 0.988 & 0.988 \\
Qwen (v\_bar, Few-shot-5) & 0.901 & 0.901 \\
DeepSeek (h\_bar, Few-shot-5) & 0.727 & 0.727 \\

\bottomrule
\end{tabular}
}

\caption{Effective attack success rate (ASR\_eff) with and without the prompt-based guard defense.}
\label{tab:guard_results}
\end{table}
\begin{table*}[t]
\centering
\tiny
\rowcolors{2}{gray!10}{white}
\renewcommand{\arraystretch}{0.85}
\setlength{\aboverulesep}{1pt}
\setlength{\belowrulesep}{1pt}
\resizebox{\textwidth}{!}{%
\begin{tabular}{lccc ccc ccc}
\rowcolor{yellow!45}
& \multicolumn{3}{c}{\textbf{PlotQA}} &
\multicolumn{3}{c}{\textbf{ChartQA}} &
\multicolumn{3}{c}{\textbf{ChartX}} \\
\textbf{Misleader}
& Base & SFT & $\Delta$
& Base & SFT & $\Delta$
& Base & SFT & $\Delta$ \\
\midrule

3D
& 23.58 & 37.74 & {+14.16}
& 24.67 & 22.16 & {-2.51}
& 18.50 & 14.54 & {-3.96} \\

Dual axis
& 36.78 & 54.02 & {+17.24}
& 28.95 & 31.58 & {+2.63}
& 52.78 & 22.22 & {-30.56} \\

Inappropriate axis range
& 13.91 & 62.69 & {+48.78}
& 16.09 & 17.43 & {+1.34}
& 39.22 & 19.61 & {-19.61} \\

Inappropriate item order
& 33.33 & 45.06 & {+11.73}
& 16.22 & 17.57 & {+1.35}
& 63.16 & 47.37 & {-15.79} \\

Inappropriate use of line
& 26.92 & 48.08 & {+21.16}
& 46.67 & 49.52 & {+2.85}
& 41.18 & 17.65 & {-23.53} \\

Inappropriate use of log scale
& 24.85 & 39.88 & {+15.03}
& 43.12 & 38.99 & {-4.13}
& -- & -- & -- \\

Inappropriate use of stacked
& 18.54 & 31.01 & {+12.47}
& 26.73 & 23.65 & {-3.08}
& 14.93 & 10.45 & {-4.48} \\

Ineffective color scheme
& 27.45 & 42.48 & {+15.03}
& 31.43 & 32.86 & {+1.43}
& 36.36 & 21.21 & {-15.15} \\

Inverted axis
& 34.17 & 53.85 & {+19.68}
& 55.90 & 41.03 & {-14.87}
& 24.81 & 21.80 & {-3.01} \\

Misrepresentation
& 39.86 & 58.36 & {+18.50}
& 37.09 & 38.74 & {+1.65}
& 6.12 & 7.35 & {+1.23} \\

Truncated axis
& 25.75 & 68.86 & {+43.11}
& 19.36 & 19.08 & {-0.28}
& 66.67 & 44.44 & {-22.23} \\

None
& 76.64 & 65.75 & {-10.89}
& 91.82 & 80.08 & {-11.74}
& 94.50 & 70.61 & {-23.89} \\

\bottomrule
\end{tabular}
}
\caption{Accuracy (\%) of the base and SFT Qwen2.5-VL-3B models on the AttackViz test set across misleaders and test subsets. $\Delta$ denotes the change
in percentage points from the base model to the SFT model.}

\label{tab:sft_results}
\end{table*}

\paragraph{Prompt-based guard.} 
We conduct a preliminary experiment by adding a system-level guard instruction to the Misleader-generator. The guard warns about adversarial distortions, forbids perceptual manipulation, and treats the misleading demonstrations as attacks that must not be followed. We provide the system-level guard prompt in Appendix \ref{sec:appendix_mitigation_guard}. In few-shot settings, the system instruction appears once, while repeated user-assistant demonstrations instantiate the prohibited behavior and output format, creating a strong conflicting in-context pattern. 

We evaluate attack success using ASR\_eff, which counts only structurally valid distortions (non-constant scaling factors and inconsistent dual-axis ranges). Table \ref{tab:guard_results} shows that across three attacker settings, the guard prompt does not reduce the attack success rate on the test set, indicating that simple prompt-level safeguards are insufficient against design-level chart attacks.

This result aligns with known limitations of instruction-following under conflicting context. In the few-shot setting, a single system instruction is outweighed by repeated demonstrations of the prohibited behavior, weakening adherence to the system-level constraint \cite{zhang-etal-2025-iheval, kim-etal-2025-really}. In the zero-shot setting, framing the attack as chart generation conceals harmful intent within an otherwise benign task, consistent with scenario nesting and intent concealment in prompt-based attacks \cite{ding-etal-2024-wolf, cui-etal-2025-exploring}.

\paragraph{Fine-tuned MLLM on AttackViz.}
We fine-tune Qwen2.5-VL-3B-Instruct on AttackViz using QLoRA \cite{dettmersQLoRA2023} with 4-bit NF4 quantization, applying LoRA adapters to the attention and feed-forward layers. Training uses PEFT \cite{peft} and TRL \cite{vonwerra2020trl}. Full training details are provided in Appendix \ref{sec:appendix_mitigation_sft}. 

We compare the fine-tuned model with its quantized instruct base model. On the in-domain PlotQA partition, the base model achieves 41.66\% accuracy on the AttackViz test set, while the fine-tuned model reaches 52.17\% (+10.51 pp). As shown in Table \ref{tab:sft_results}, on the PlotQA partition performance improves across all misleaders, with gains ranging from 11.73 to 48.78 percentage points.

Cross-domain transfer is mixed: on ChartQA, six misleaders improve (+1.34 to +2.85 pp), while the others decline, most notably inverted axis (-14.87 pp); on ChartX, nine of ten applicable misleaders decline, with only \textit{misrepresentation} improving (+1.23 pp). Clean-chart accuracy also decreases by 10.89 pp on PlotQA, 11.74 pp on ChartQA, and 23.89 pp on ChartX. A similar trade-off between robustness to misleading charts and clean-chart performance was observed by \citet{tonglet2025protecting} for inference-time mitigation. These preliminary results show that robustness-oriented fine-tuning can substantially improve in-domain robustness, but its limited cross-domain transfer and degradation on clean charts indicate that the gains do not yet generalize reliably.

\section{Conclusions}
We present a systematic study of how MLLMs can generate misleading charts. We introduce ChartAttack, an automated framework for applying design-level misleaders to chart annotations, and show that the resulting charts substantially degrade chart QA performance across models and datasets. Importantly, these errors reflect targeted deception rather than only general performance degradation: originally correct answers frequently shift to attacker-intended answers, while originally incorrect answers rarely change. Stronger chart-reasoning models are not immune, and vulnerability varies across model families and misleaders. A complementary human study further shows that these attacks impair human chart comprehension, demonstrating that the observed effects extend beyond MLLM evaluation. To support further research, we release AttackViz, a dataset of paired correct and misleading charts annotated with applied misleaders and their induced incorrect answers. Fine-tuning on AttackViz improves in-domain robustness, but limited cross-domain transfer and reduced clean-chart performance indicate that robust mitigation remains an open challenge. Overall, our findings expose an underexplored attack surface in multimodal chart generation and show that data faithfulness alone is insufficient: trustworthy systems must also be robust to adversarial visualization design and targeted perceptual manipulation.

\newpage

\section*{Limitations}

We identify four limitations in this work.

First, our framework and dataset focus on three chart types. However, this selection covers a large share of real-world cases: 64\% of the misleading charts in the taxonomy proposed by \cite{lo2022misinformed} and 49\% of those in the MisViz benchmark \cite{tonglet2025chartlyingmeautomating}.

Second, our study focuses on a subset of misleader categories, specifically design misleaders \citep{lo2022misinformed}. Reasoning misleaders, which manipulate titles or annotations without violating explicit design rules, remain underexplored. By limiting the scope, we maintain controlled evaluation of misleader effects while acknowledging that our dataset does not cover all possible real-world misleader scenarios.

Third, both AttackViz and ChartAttack inherit a dependence on the models used to construct them. AttackViz was built through a model-in-the-loop filtering process that keeps correct charts answerable while ensuring misleading variants induce incorrect interpretations, enabling controlled evaluation of misleader effects; a consequence is that the dataset may over-represent patterns effective against the filtering models. Relatedly, because the Misleader-Generator uses code-instruction models, the generated attacks may reflect generator-specific preferences. These dependencies have limited effect on our findings, as the misleading charts remain effective on a separate set of more recent MLLM families and are validated by human participants, indicating that AttackViz generalizes beyond its original model set and remains a useful diagnostic for studying misleader effects and evaluating defenses.

Fourth, our human evaluation was conducted with a relatively limited participant population and focused on chart QA accuracy in a controlled setting. This setting was chosen to isolate the effect of visualization-level misleaders while avoiding confounds from surrounding text, source credibility, layout, interactivity, topic familiarity, and users’ prior beliefs. Future work should evaluate broader populations, additional chart-analysis tasks, and richer communication contexts such as news articles, dashboards, and social media.

\section*{Ethics statement}

This work examines how MLLMs may be misused to generate misleading charts at scale, with the goal of raising awareness of this risk and motivating stronger robustness and security considerations in chart generation systems. Understanding how MLLMs could be exploited to generate misleading charts is essential for designing effective defenses. Our work analyzes potential attacks not to promote misuse, but to inform robust detection, mitigation, and responsible visualization practices. While such techniques could be exploited to manipulate information, we follow principles of responsible disclosure by providing sufficient detail to support analysis, detection, and mitigation.

\paragraph{Human study.} The human evaluation was conducted as an exploratory study with informed consent, without collecting any personal data, and all responses were anonymous. No harm to individuals or organizations occurred during the study. We encourage future work to build on these findings to develop detection methods, robustness-aware training, and safeguards that promote trustworthy data communication in real-world visualization tools.

\paragraph{Dataset access.} 
Our code is released under the Apache 2.0 license. Our dataset combines annotations from PlotQA \citep{methani2020plotqa} (CC BY 4.0), ChartQA \citep{masry2022chartqa} (GPLv3) and ChartX \citep{xia2025chartx}. Because ChartQA is GPLv3, the combined dataset is released under GPLv3.

\paragraph{AI assistants use.} We use AI assistants in this work to help with writing by correcting grammar mistakes and typos.

\section*{Acknowledgments}
This research was partially funded by the Ministry of Education and Science of Bulgaria (support for INSAIT, part of the Bulgarian National Roadmap for Research Infrastructure), the German Federal Ministry of Research, Technology and Space and the Hessian Ministry of Higher Education, Research, Science and the Arts within their joint support of the National Research Center for Applied Cybersecurity ATHENE, and by the LOEWE Distinguished Chair ``Ubiquitous Knowledge Processing'', LOEWE initiative, Hesse, Germany (Grant Number: LOEWE/4a//519/05/00.002(0002)/81). We thank Federico Marcuzzi, Shivam Sharma and Hassan Soliman for their feedback on an early draft of this work.

\bibliography{custom}

\appendix
\section{Demonstration selection module: Training dataset creation}
\label{sec:appendix_demonstration_selection_training_dataset}
The first step in fine-tuning the Demonstration Selection module of ChartAttack is to create a suitable dataset. We use the training split of AttackViz for this purpose. Using the chart JSON annotations and the set of misleaders that affect each chart, we construct anchor-positive pairs. A pair is considered similar if the sets of misleaders match exactly (Jaccard index = 1). To reduce the length of input sequences, we apply an annotation simplification step. We remove most display and styling metadata, including titles, legends, grids, font sizes, labels, horizontal bands, and chart legends, keeping only core data, axes, colors, chart type, and basic chart settings such as stacking and 3D effects. We also remove JSON-specific characters. Each pair is represented by concatenating the question with the simplified chart annotation JSON.

\section{ChartAttack: Ablation experiments}
\label{sec:appendix_ablations}
\subsection{Demonstration selection module}
\begin{table*}[!ht]
\centering
\footnotesize
\begin{tabularx}{\textwidth}{
    p{3.0cm}
    p{1.0cm}
    p{3.0cm}
    >{\centering\arraybackslash}X
    >{\centering\arraybackslash}X
    >{\centering\arraybackslash}X
}
\rowcolor{yellow!45}
\textbf{Model} & \textbf{Loss} & \textbf{Downsampling} &
\textbf{Horizontal bar} & \textbf{Vertical bar} & \textbf{Line} \\
\midrule
\texttt{BM25} & -- & anchor-positive
 & 40.56 & 34.45 & 78.72 \\
\hline
\multirow{2}{*}{\texttt{all-mpnet-base-v2}}
 & MNR & anchor
 & 45.28 & 42.15 & \digitbf{80.85} \\
 & MNR & anchor-positive
 & \digitbf{46.17} & \digitbf{42.54} & 80.14 \\
\hline
\multirow{2}{*}{\footnotesize\shortstack[l]{mxbai-embed-large-v1/\\all-mpnet-base-v2}}
 & GISTE & anchor
 & 42.35 & 39.07 & 78.72 \\
 & GISTE & anchor-positive
 & 39.33 & 39.33 & 79.43 \\
\end{tabularx}
\caption{Accuracy@5 on the validation set of AttackViz under different objectives and downsampling strategies.  Best results are marked in \digitbf{bold}.}
\label{tab:demonstration_module}
\end{table*}

\label{subsubsec:demonstration_ablation}
We evaluate the demonstration selection module using MNR \citep{henderson2017efficient} and GISTE \citep{solatorio2024gistembed} losses with median-based downsampling to balance the anchor-positive dataset from AttackViz. For each instance, we compute a maximum allowable frequency $t=(median / mean) \times median$ and downsample instances exceeding it, applying the strategy either to anchor texts alone or to both anchor and positive texts. We also compare against lexical BM25 \citep{robertson1995okapi}. Table \ref{tab:demonstration_module} reports Accuracy@5 on the validation split. SBERT with MNR and downsampling on anchor-positive texts achieves the highest Accuracy@5 for horizontal bar (46.17) and vertical bar (42.54), while anchor-only MNR performs best on line charts (80.85). GISTE shows mixed results, slightly lowering bar chart scores but maintaining line chart accuracy. BM25 performs worst.

We perform oracle experiments to choose the number of demonstrations for the misleader-generator module of ChartAttack. We report results on the validation split of AttackViz. Similar to the ablation used to select the few-shot strategy, we frame this task as a multi-label classification problem, where a chart JSON annotation–question pair may have one or more misleaders. Table \ref{tab:oracle_results} reports results for one-, three-, and five-shot prompting. Moving from one to three shots yields large performance gains across all chart types, with Macro F1 improving from 0.39-0.52 in the 1-shot setting to 0.55-0.92 in the 3-shot setting. Increasing the number of shots from three to five results in smaller but consistent improvements, with Macro F1 rising by up to 0.05 for horizontal bar charts and by 0.13 for line charts. The five-shot setting achieves the highest Macro F1-score for all chart types, indicating improved balance across misleader categories rather than gains driven by dominant labels. Based on these quantitative improvements, we adopt five demonstrations in our final configuration as a practical trade-off between performance and prompt length.

\begin{table*}[!ht]
\centering
\footnotesize
\resizebox{\textwidth}{!}{%
\begin{tabular}{l c c c c c c c c c c}
\rowcolor{yellow!45} 
\textbf{Chart type} & \textbf{Dual axis} & \textbf{Inverted axis} & \textbf{Log scale} & \textbf{Line} & \textbf{Stacked} & \textbf{3D} & \textbf{Color} & \textbf{Misrepresentation} & \textbf{Micro F1} & \textbf{Macro F1} \\
\midrule
\rowcolor{gray!20}
\multicolumn{11}{c}{\textbf{1-shot}} \\
Horizontal bar & 0.62 & 0.59 & 0.65 & - & 0.26 & 0.48 & 0.53 & 0.53 & 0.48 & 0.52 \\
Vertical bar & 0.31 & 0.68 & 0.54 & 0.25 & 0.35 & 0.34 & 0.76 & 0.58 & 0.48 & 0.48 \\
Line & 0.07 & 0.34 & 0.67 & - & - & - & - & 0.46 & 0.41 & 0.39 \\
\midrule
\rowcolor{gray!20}
\multicolumn{11}{c}{\textbf{3-shot}} \\
Horizontal bar & 0.61 & 0.87 & 0.87 & - & 0.79 & 0.96 & 0.94 & 0.92 & 0.86 & 0.85 \\
Vertical bar & 1 & 0.97 & 0.88 & 0.9 & 0.86 & 0.96 & 0.9 & 0.89 & 0.9 & \digitbf{0.92} \\
Line & 0 & 0.89 & 0.77 & - & - & - & - & 0.54 & 0.75 & 0.55 \\
\midrule
\rowcolor{gray!20}
\multicolumn{11}{c}{\textbf{5-shot}} \\
Horizontal bar & 0.77 & 0.98 & 0.88 & - & 0.82 & 0.96 & 0.94 & 0.95 & 0.89 & \digitbf{0.9} \\
Vertical bar & 0 & 0.96 & 0.96 & 0.96 & 0.92 & 0.97 & 1 & 0.94 & 0.95 & 0.84 \\
Line & 0 & 0.91 & 0.85 & - & - & - & - & 0.95 & 0.9 & \digitbf{0.68}\\
\end{tabular}
}
\caption{Oracle experiment results by chart type and misleading technique across different few-shot settings. Best results are marked in \digitbf{bold}. "-" indicates that a misleader is not applicable to a specific chart type.}
\label{tab:oracle_results}
\end{table*}

\subsection{Misleader-generator module}
We compare eight open-weight, instruction-tuned code models from three families: DeepSeek-Coder \citep{guo2024deepseek}, Qwen 2.5-Coder \citep{hui2024qwen25codertechnicalreport}, and Qwen 3.0-Coder \citep{yang2025qwen3}, with 1.3B-33B parameters. These are code-instruction models, whereas the models evaluated in Section \ref{subsection:experimental_setup} are instruction-tuned MLLMs from diverse families \cite{ling2025Beyond, zhao2025ChartCoder}. Prior work shows that specialized variants can differ even within the same model family, and that code-specialized backbones and training choices can substantially affect chart-to-code performance. We evaluate zero-shot, random 5-shot, and demonstration 5-shot prompting using our Demonstration Selection module, where random 5-shot selects same–chart-type instances per query from the AttackViz training split. We frame the task as multi-label classification and report Macro-F1 on the AttackViz validation split, where each chart annotation-question pair may contain multiple misleaders.

\begin{figure}
    \centering
    \includegraphics[width=\linewidth]{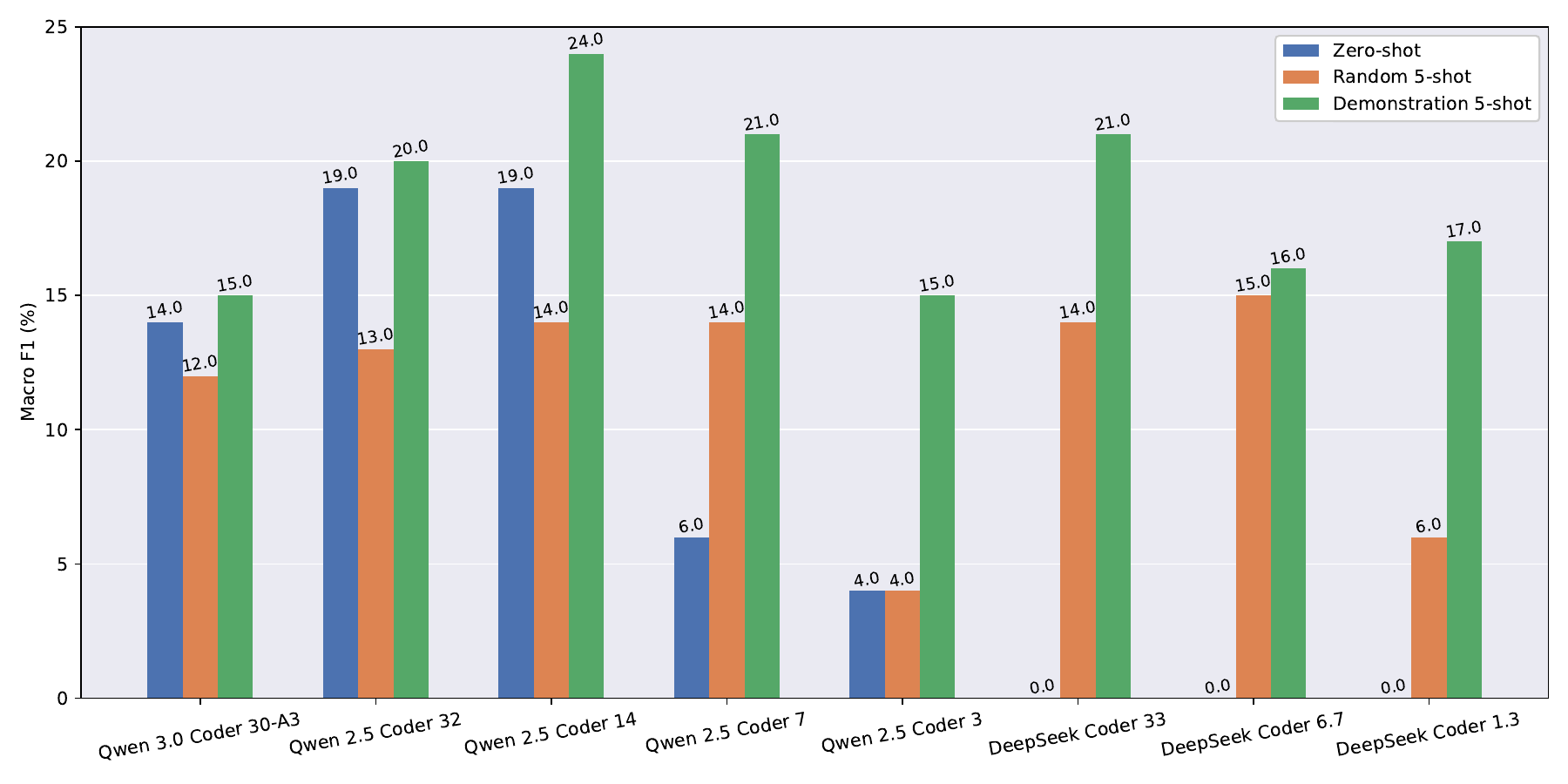}
    \caption{Average Macro F1 of the eight code models evaluated as Misleader-generator module. Few-shot strategies are encoded by color: \textbf{zero-shot} (\zeroshotmarker), \textbf{random few-shot} (\randomshotmarker), and \textbf{demonstration few-shot} (\demoshotmarker).}
    \label{fig:ablation_attacker}
\end{figure}

Figure \ref{fig:ablation_attacker} shows average results by model. Zero-shot performance varies widely: Qwen models achieve moderate scores, while DeepSeek models fail. Random 5-shot provides limited gains for weaker models and can hurt strong zero-shot models. Demonstration 5-shot performs best across all models, making zero-shot–weak models competitive and often allowing smaller models to outperform larger ones. In ChartAttack, we select attackers and prompting strategies by chart type: Qwen-Coder 14B with demonstration 5-shot for vertical bar charts, Qwen-Coder 14B with zero-shot for line charts, and DeepSeek-Coder 33B with demonstration 5-shot for horizontal bar charts.

\section{AttackViz corpus}
\label{sec:appendix_attackViz corpus}

\subsection{Misleader selection}
\label{sec:appendix_attackViz_misleader}

Table \ref{tab:misleaders_properties} shows all the misleaders proposed in the taxonomy of \citep{lo2022misinformed}. Each column corresponds to one of the criteria used to select the final subset of misleaders in this work. The criteria are the following:

\begin{itemize}
    \item \textbf{Correct answer unchanged}: The misleader does not affect the correct answer to a question associated with the chart.
    
    \item \textbf{Violates chart grammar}: These misleaders break visualization design principles that may lead to incorrect conclusions about the underlying data.
    
    \item \textbf{Data unchanged}: These misleaders do not modify the underlying data table used to generate the chart; therefore, the correct conclusion can still be reached.
    
    \item \textbf{Python implementable}: The misleader can be implemented in Python using Matplotlib.
    
    \item \textbf{$5+$ occurrences}: These misleaders appear frequently in real-world examples.
    
    \item \textbf{Previously studied}: These misleaders have been previously studied in misleading chart QA \citep{ge2023calvi,bharti2024chartom} or design-support research \citep{lo2023change}
\end{itemize}

\begin{table*}[t]
\centering
\scriptsize
\setlength{\tabcolsep}{4pt}
\renewcommand{\arraystretch}{1.00}

\rowcolors{2}{gray!10}{white}

\begin{tabularx}{\textwidth}{
>{\raggedright\arraybackslash}p{4.2cm}
*{6}{>{\centering\arraybackslash}X}
}

\rowcolor{yellow!45}
\textbf{Misleader} &
\textbf{\shortstack{Correct answer\\unchanged}} &
\textbf{\shortstack{Violates chart\\grammar}} &
\textbf{\shortstack{Data\\unchanged}} &
\textbf{\shortstack{Python\\implementable}} &
\textbf{\shortstack{$5+$\\occurrences}} &
\textbf{\shortstack{Previously\\studied}} \\

\midrule

Not data &  &\includegraphics[height=0.8em]{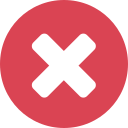}&\includegraphics[height=0.8em]{images/icons/crossmark.png}&\includegraphics[height=0.8em]{images/icons/crossmark.png}&  &\includegraphics[height=0.8em]{images/icons/crossmark.png}\\
Selective data &\includegraphics[height=0.8em]{images/icons/crossmark.png}&\includegraphics[height=0.8em]{images/icons/crossmark.png}&  &  &  &  \\
Dubious data &  &\includegraphics[height=0.8em]{images/icons/crossmark.png}&\includegraphics[height=0.8em]{images/icons/crossmark.png}&  &  &\includegraphics[height=0.8em]{images/icons/crossmark.png}\\
Non sequitur &  &\includegraphics[height=0.8em]{images/icons/crossmark.png}&\includegraphics[height=0.8em]{images/icons/crossmark.png}&  &  &\includegraphics[height=0.8em]{images/icons/crossmark.png}\\
Too few data points &\includegraphics[height=0.8em]{images/icons/crossmark.png}&\includegraphics[height=0.8em]{images/icons/crossmark.png}&  &  &  &\includegraphics[height=0.8em]{images/icons/crossmark.png}\\
Discretized continuous variable &\includegraphics[height=0.8em]{images/icons/crossmark.png}&  &  &  &  &\includegraphics[height=0.8em]{images/icons/crossmark.png}\\
Missing normalization &  &\includegraphics[height=0.8em]{images/icons/crossmark.png}&  &  &  &  \\
\digitbf{Inappropriate item order} &  &  &  &  &  &  \\
Inappropriate metric &  &\includegraphics[height=0.8em]{images/icons/crossmark.png}&  &  &  &\includegraphics[height=0.8em]{images/icons/crossmark.png}\\
Questionable prediction &  &\includegraphics[height=0.8em]{images/icons/crossmark.png}&  &  &  &\includegraphics[height=0.8em]{images/icons/crossmark.png}\\
Trend line on random data &  &\includegraphics[height=0.8em]{images/icons/crossmark.png}&\includegraphics[height=0.8em]{images/icons/crossmark.png}&  &  &\includegraphics[height=0.8em]{images/icons/crossmark.png}\\
Inappropriate use of accumulation &\includegraphics[height=0.8em]{images/icons/crossmark.png}&  &  &  &  &\includegraphics[height=0.8em]{images/icons/crossmark.png}\\
Inappropriate aggregation granularity &  &\includegraphics[height=0.8em]{images/icons/crossmark.png}&  &  &\includegraphics[height=0.8em]{images/icons/crossmark.png}&  \\
Two-way normalization &  &\includegraphics[height=0.8em]{images/icons/crossmark.png}&  &  &\includegraphics[height=0.8em]{images/icons/crossmark.png}&\includegraphics[height=0.8em]{images/icons/crossmark.png}\\
\digitbf{Truncated axis} & &  &  &  &  &  \\
\digitbf{Dual axis} &  &  &  &  &  &  \\
\digitbf{Inappropriate axis range} &  &  &  &  &  &  \\
\digitbf{Inverted axis} &  &  &  &  &  &  \\
\digitbf{Log scale} &  &  &  &  &  &  \\
Extended axis &  &  &  &\includegraphics[height=0.8em]{images/icons/crossmark.png}&  &\includegraphics[height=0.8em]{images/icons/crossmark.png}\\
Data of different magnitudes &  &\includegraphics[height=0.8em]{images/icons/crossmark.png}&  &  &  &\includegraphics[height=0.8em]{images/icons/crossmark.png}\\
Linear scale on exponential data &  &  &  &\includegraphics[height=0.8em]{images/icons/crossmark.png}&  &\includegraphics[height=0.8em]{images/icons/crossmark.png}\\
\digitbf{Inappropriate use of line chart} &  &  &  &  &  &  \\
Inappropriate use of pie chart &\includegraphics[height=0.8em]{images/icons/crossmark.png}&  &  &  &  &\includegraphics[height=0.8em]{images/icons/crossmark.png}\\
Confusing chart type &  &\includegraphics[height=0.8em]{images/icons/crossmark.png}&  &\includegraphics[height=0.8em]{images/icons/crossmark.png}&  &\includegraphics[height=0.8em]{images/icons/crossmark.png}\\
Misusing circular layout &  &  &  &\includegraphics[height=0.8em]{images/icons/crossmark.png}&  &\includegraphics[height=0.8em]{images/icons/crossmark.png}\\
\digitbf{Inappropriate use of stacked} &  &  &  &  &  &  \\
Inappropriate use of bar chart &  &  &  &\includegraphics[height=0.8em]{images/icons/crossmark.png}&  &\includegraphics[height=0.8em]{images/icons/crossmark.png}\\
Inappropriate use of scatterplot &  &  &  &\includegraphics[height=0.8em]{images/icons/crossmark.png}&  &\includegraphics[height=0.8em]{images/icons/crossmark.png}\\
\digitbf{Overusing colors} &  &  &  &  &  &  \\
\digitbf{Indistinguishable colors} &  &  &  &  &  &  \\
Color blind unfriendly &  &  &  &\includegraphics[height=0.8em]{images/icons/crossmark.png}&  &\includegraphics[height=0.8em]{images/icons/crossmark.png}\\
Missing title &\includegraphics[height=0.8em]{images/icons/crossmark.png}&\includegraphics[height=0.8em]{images/icons/crossmark.png}&  &  &  &\includegraphics[height=0.8em]{images/icons/crossmark.png}\\
Missing axis title &\includegraphics[height=0.8em]{images/icons/crossmark.png}&\includegraphics[height=0.8em]{images/icons/crossmark.png}&  &  &  &\includegraphics[height=0.8em]{images/icons/crossmark.png}\\
Missing legend &\includegraphics[height=0.8em]{images/icons/crossmark.png}&\includegraphics[height=0.8em]{images/icons/crossmark.png}&  &  &  &\includegraphics[height=0.8em]{images/icons/crossmark.png}\\
Missing value labels &\includegraphics[height=0.8em]{images/icons/crossmark.png}&\includegraphics[height=0.8em]{images/icons/crossmark.png}&  &  &  &\includegraphics[height=0.8em]{images/icons/crossmark.png}\\
Missing axis &\includegraphics[height=0.8em]{images/icons/crossmark.png}&\includegraphics[height=0.8em]{images/icons/crossmark.png}&  &  &  &\includegraphics[height=0.8em]{images/icons/crossmark.png}\\
Missing axis ticks &\includegraphics[height=0.8em]{images/icons/crossmark.png}&\includegraphics[height=0.8em]{images/icons/crossmark.png}&  &  &  &\includegraphics[height=0.8em]{images/icons/crossmark.png}\\
Missing units &\includegraphics[height=0.8em]{images/icons/crossmark.png}&\includegraphics[height=0.8em]{images/icons/crossmark.png}&  &  &\includegraphics[height=0.8em]{images/icons/crossmark.png}&\includegraphics[height=0.8em]{images/icons/crossmark.png}\\
\digitbf{Misrepresentation} &  &  &  &  &  &  \\
Inconsistent tick intervals &\includegraphics[height=0.8em]{images/icons/crossmark.png}&  &  &  &  &\includegraphics[height=0.8em]{images/icons/crossmark.png}\\
Inconsistent binning size &\includegraphics[height=0.8em]{images/icons/crossmark.png}&  &  &  &  &\includegraphics[height=0.8em]{images/icons/crossmark.png}\\
Changing scale &\includegraphics[height=0.8em]{images/icons/crossmark.png}&  &  &  &  &\includegraphics[height=0.8em]{images/icons/crossmark.png}\\
Violating color convention &  &\includegraphics[height=0.8em]{images/icons/crossmark.png}&  &  &\includegraphics[height=0.8em]{images/icons/crossmark.png}&\includegraphics[height=0.8em]{images/icons/crossmark.png}\\
Inconsistent grouping &\includegraphics[height=0.8em]{images/icons/crossmark.png}&\includegraphics[height=0.8em]{images/icons/crossmark.png}&  &  &\includegraphics[height=0.8em]{images/icons/crossmark.png}&\includegraphics[height=0.8em]{images/icons/crossmark.png}\\
Inconsistent tick labels &  &\includegraphics[height=0.8em]{images/icons/crossmark.png}&  &  &\includegraphics[height=0.8em]{images/icons/crossmark.png}&\includegraphics[height=0.8em]{images/icons/crossmark.png}\\
Inconsistent value labels &  &  &  &\includegraphics[height=0.8em]{images/icons/crossmark.png}&  &\includegraphics[height=0.8em]{images/icons/crossmark.png}\\
Cluttering &\includegraphics[height=0.8em]{images/icons/crossmark.png}&  &  &  &  &\includegraphics[height=0.8em]{images/icons/crossmark.png}\\
Confusing legend &\includegraphics[height=0.8em]{images/icons/crossmark.png}&\includegraphics[height=0.8em]{images/icons/crossmark.png}&  &  &  &\includegraphics[height=0.8em]{images/icons/crossmark.png}\\
Plotting error &  &\includegraphics[height=0.8em]{images/icons/crossmark.png}&  &\includegraphics[height=0.8em]{images/icons/crossmark.png}&  &\includegraphics[height=0.8em]{images/icons/crossmark.png}\\
Missing abbreviation &  &\includegraphics[height=0.8em]{images/icons/crossmark.png}&  &  &  &\includegraphics[height=0.8em]{images/icons/crossmark.png}\\
Misalignment &  &  &  &\includegraphics[height=0.8em]{images/icons/crossmark.png}&  &\includegraphics[height=0.8em]{images/icons/crossmark.png}\\
Plotting out of chart &\includegraphics[height=0.8em]{images/icons/crossmark.png}&  &  &  &  &\includegraphics[height=0.8em]{images/icons/crossmark.png}\\
Illegible text &\includegraphics[height=0.8em]{images/icons/crossmark.png}&\includegraphics[height=0.8em]{images/icons/crossmark.png}&  &  &  &\includegraphics[height=0.8em]{images/icons/crossmark.png}\\
\digitbf{3D} &  &  &  &  &  &  \\
Area encoding &\includegraphics[height=0.8em]{images/icons/crossmark.png}&  &  &\includegraphics[height=0.8em]{images/icons/crossmark.png}&  &\includegraphics[height=0.8em]{images/icons/crossmark.png}\\
\digitbf{Ineffective color scheme} &  &  &  &  &  &  \\
Pictorial area encoding &  &  &  &\includegraphics[height=0.8em]{images/icons/crossmark.png}&  &  \\
Inappropriate use of smoothing &  &\includegraphics[height=0.8em]{images/icons/crossmark.png}&  &  &\includegraphics[height=0.8em]{images/icons/crossmark.png}&\includegraphics[height=0.8em]{images/icons/crossmark.png}\\
Distractive value labels &  &\includegraphics[height=0.8em]{images/icons/crossmark.png}&  &  &\includegraphics[height=0.8em]{images/icons/crossmark.png}&\includegraphics[height=0.8em]{images/icons/crossmark.png}\\
Map projection distortion &\includegraphics[height=0.8em]{images/icons/crossmark.png}&  &  &  &\includegraphics[height=0.8em]{images/icons/crossmark.png}&\includegraphics[height=0.8em]{images/icons/crossmark.png}\\
Inappropriate aspect ratio &  &  &  &\includegraphics[height=0.8em]{images/icons/crossmark.png}&  &\includegraphics[height=0.8em]{images/icons/crossmark.png}\\
Sine illusion &  &\includegraphics[height=0.8em]{images/icons/crossmark.png}&  &  &\includegraphics[height=0.8em]{images/icons/crossmark.png}&\includegraphics[height=0.8em]{images/icons/crossmark.png}\\
Invalid comparison &  &\includegraphics[height=0.8em]{images/icons/crossmark.png}&  &  &  &\includegraphics[height=0.8em]{images/icons/crossmark.png}\\
Correlation not causation &  &\includegraphics[height=0.8em]{images/icons/crossmark.png}&  &  &  &\includegraphics[height=0.8em]{images/icons/crossmark.png}\\
Pattern seeking &  &\includegraphics[height=0.8em]{images/icons/crossmark.png}&  &  &  &\includegraphics[height=0.8em]{images/icons/crossmark.png}\\
Misleading claim &\includegraphics[height=0.8em]{images/icons/crossmark.png}&\includegraphics[height=0.8em]{images/icons/crossmark.png}&  &  &  &\includegraphics[height=0.8em]{images/icons/crossmark.png}\\
Misleading annotation &\includegraphics[height=0.8em]{images/icons/crossmark.png}&\includegraphics[height=0.8em]{images/icons/crossmark.png}&  &  &  &  \\
Misleading title &\includegraphics[height=0.8em]{images/icons/crossmark.png}&\includegraphics[height=0.8em]{images/icons/crossmark.png}&  &  &\includegraphics[height=0.8em]{images/icons/crossmark.png}&  \\
Misleading value labels &\includegraphics[height=0.8em]{images/icons/crossmark.png}&\includegraphics[height=0.8em]{images/icons/crossmark.png}&  &  &\includegraphics[height=0.8em]{images/icons/crossmark.png}&  \\
Hidden distribution &  &\includegraphics[height=0.8em]{images/icons/crossmark.png}&  &  &  &\includegraphics[height=0.8em]{images/icons/crossmark.png}\\
Overplotting &\includegraphics[height=0.8em]{images/icons/crossmark.png}&  &  &  &  &  \\
Hidden uncertainty &\includegraphics[height=0.8em]{images/icons/crossmark.png}&\includegraphics[height=0.8em]{images/icons/crossmark.png}&  &  &\includegraphics[height=0.8em]{images/icons/crossmark.png}&  \\
Hidden population size &\includegraphics[height=0.8em]{images/icons/crossmark.png}&\includegraphics[height=0.8em]{images/icons/crossmark.png}&  &  &\includegraphics[height=0.8em]{images/icons/crossmark.png}&\includegraphics[height=0.8em]{images/icons/crossmark.png}\\

\bottomrule
\end{tabularx}

\caption{Properties of the misleaders. \includegraphics[height=0.8em]{images/icons/crossmark.png} indicates that the corresponding property does not apply. Misleaders satisfying all criteria are highlighted in \digitbf{bold}.}
\label{tab:misleaders_properties}
\end{table*}

As shown in Table \ref{tab:misleaders_properties}, 13 out of 74 misleaders satisfy all the considered criteria. Moreover, there is substantial overlap between Overusing colors, Indistinguishable colors, and Ineffective color scheme. As a result, we merge these misleaders into a single category, Ineffective color scheme, resulting in a final set of 11 misleaders.

\subsection{Cross-domain extension}
\label{sec:appendix_attackViz_cross}
Table \ref{tab:original_chartqa} shows the statistics of the ChartQA dataset as reported by \citet{masry2022chartqa}. The most significant reduction in dataset size is due to incomplete Chart JSON annotations and missing CSV table data, which prevent the reconstruction of charts or omit essential visual encoding information such as bar or line colors. Because AttackViz aims to generate synthetic charts that closely resemble real-world charts, we discard such incomplete instances. For all experiments involving ChartQA, we merge all dataset partitions and use the resulting set exclusively for testing.
\begin{table}[H]
\centering
\setlength{\tabcolsep}{3pt} 

\begin{tabular}{c c c} 
\rowcolor{yellow!45}
\textbf{Split} & \textbf{Charts} & \textbf{Questions}\\
\midrule
Train & 19173 & 28299 \\
Validation & 1160 & 1920 \\
Test & 1612 & 2500 \\
\end{tabular}

\caption{Statistics of ChartQA by split reported by \citet{masry2022chartqa}.}
\label{tab:original_chartqa}
\end{table}

Table \ref{tab:original_chartX} shows the statistics of the ChartX evaluation set. We consider only bar and line charts based on the criteria described in Section \ref{sec:chartAttack} and Appendix \ref{sec:appendix_attackViz corpus}. We obtain the chart JSON annotations by extracting the underlying data from the CSV files provided in the dataset.

\begin{table}[H]
\centering
\setlength{\tabcolsep}{3pt} 

\begin{tabular}{c c} 
\rowcolor{yellow!45}
\textbf{Chart Type} & \textbf{Count}\\
\midrule
v\_bar & 1224 \\
line & 944 \\
\end{tabular}

\caption{Distribution of chart types in ChartX.}
\label{tab:original_chartX}
\end{table}

\subsection{AttackViz corpus: statistics}
\label{sec:appendix_attackViz_statistics}
Table~\ref{tab:corpus_statistics} summarizes the statistics of the AttackViz corpus across the train, validation, and test splits. The table reports the number of question-chart pairs for each chart type, along with the distribution of misleaders applied to the charts. A dash (-) indicates that a given misleader is not applicable to the corresponding chart type.
\begin{table*}[!ht]
\centering
\footnotesize
\resizebox{\textwidth}{!}{%
\begin{tabular}{l c c c c c c c c c c c c c}
\rowcolor{yellow!45} 
\textbf{Chart type} & \textbf{\#Q} & \textbf{Dual axis} & \textbf{Inverted axis} & \textbf{Log scale} & \textbf{Line} & \textbf{Stacked} & \textbf{3D} & \textbf{Color} & \textbf{Misrepresentation} & \textbf{Truncated axis} & \textbf{Axis range} & \textbf{Item order} \\
\midrule
\rowcolor{purple!20}
\multicolumn{13}{c}{\textbf{PlotQA \citep{methani2020plotqa}}} \\
\midrule
\rowcolor{gray!20}
\multicolumn{13}{c}{\textbf{Train}} \\
Horizontal bar & 776 & 40 & 147 & 106 & - & 470 & 229 & 106 & 174 & 100 & 235 & 37\\
Vertical bar & 788 & 18 & 123 & 139 & 182 & 424 & 198 & 136 & 169 & 67 & 180 & 21 \\
Line & 461 & 3 & 286 & 37 & - & - & - & - & 224 & - & 53 & 89 & \\
\midrule
\rowcolor{gray!20}
\multicolumn{13}{c}{\textbf{Validation}} \\
Horizontal bar & 809 & 57 & 133 & 169 & - & 476 & 212 & 97 & 174 & 96 & 214 & 40 \\
Vertical bar & 812 & 15 & 148 & 123 & 210 & 429 & 213 & 74 & 199 & 68 & 186 & 28 \\
Line & 425 & 5 & 278 & 0 & - & - & - & - & 192 & - & 49 & 83  \\
\midrule
\rowcolor{gray!20}
\multicolumn{13}{c}{\textbf{Test}} \\
Horizontal bar & 784 & 54 & 127 & 127 & - & 477 & 206 & 84 & 151 & 88 & 209 & 45 \\
Vertical bar & 787 & 22 & 147 & 165 & 156 & 413 & 218 & 69 & 193 & 79 & 196 & 30  \\
Line & 436 & 11 & 285 & - & - & - & - & - & 218 & - & 48 & 87
\end{tabular}%
}
\caption{Statistics of the AttackViz corpus by chart type and misleading technique. \#Q denotes the number of questions. Log scale, Line, Stacked, Axis range, Item order, and Color correspond to Inappropriate use of log scale, Inappropriate use of line, Inappropriate use of stacked, Inappropriate axis range, Inappropriate item order, and Ineffective color scheme, respectively.}

\label{tab:corpus_statistics}
\end{table*}

\begin{figure*}[ht]
    \centering
    \includegraphics[width=\linewidth,height=0.90\textheight,keepaspectratio]{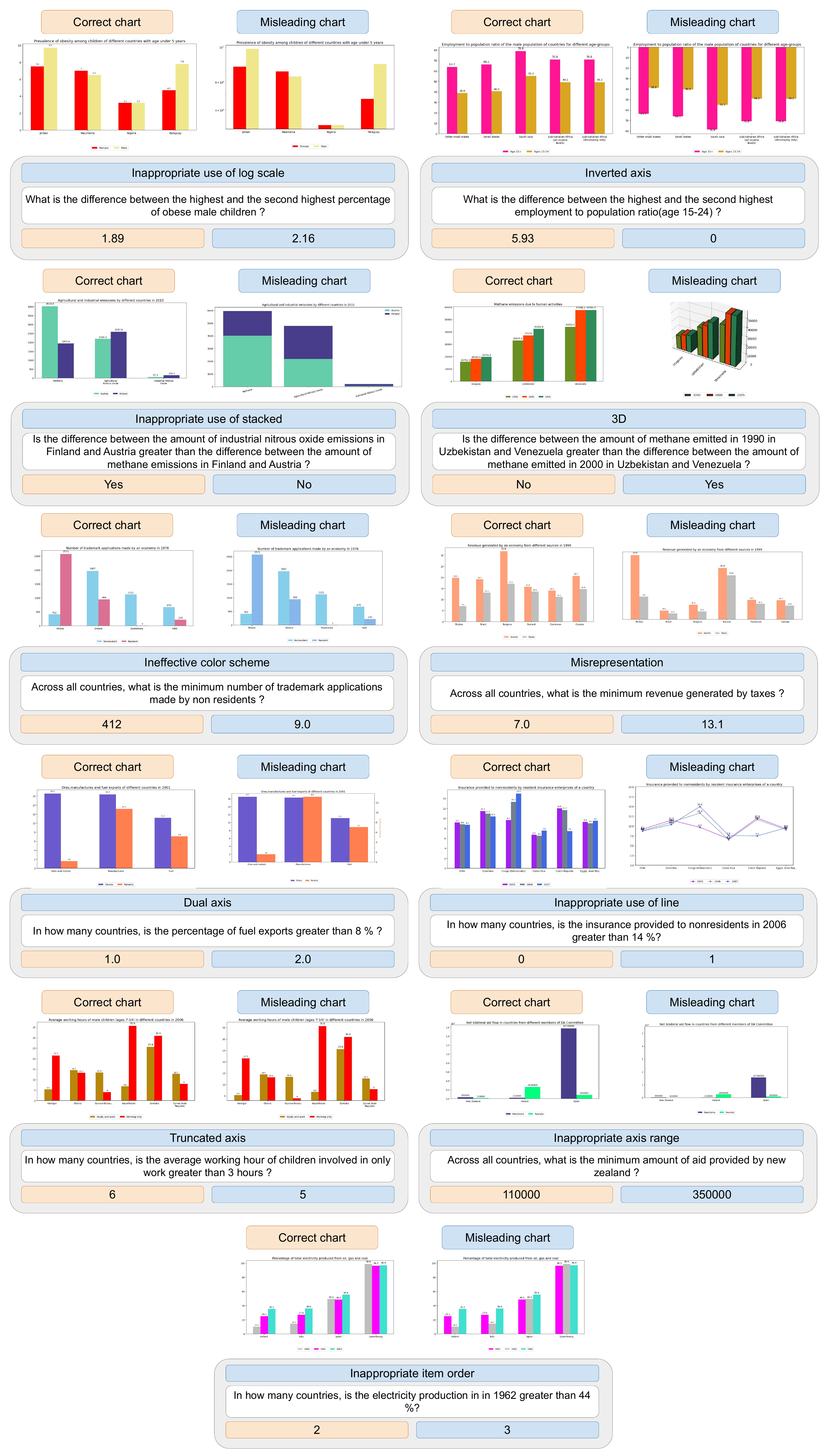}
    \caption{Examples of vertical bar charts from AttackViz. Each example includes a correct and a misleading chart, a question about the chart, and corresponding correct and misleading answers caused by the indicated misleader.}
    \label{fig:attackViz_v_bar}
\end{figure*}

\begin{figure*}[ht]
    \includegraphics[width=\linewidth,height=0.90\textheight,keepaspectratio]{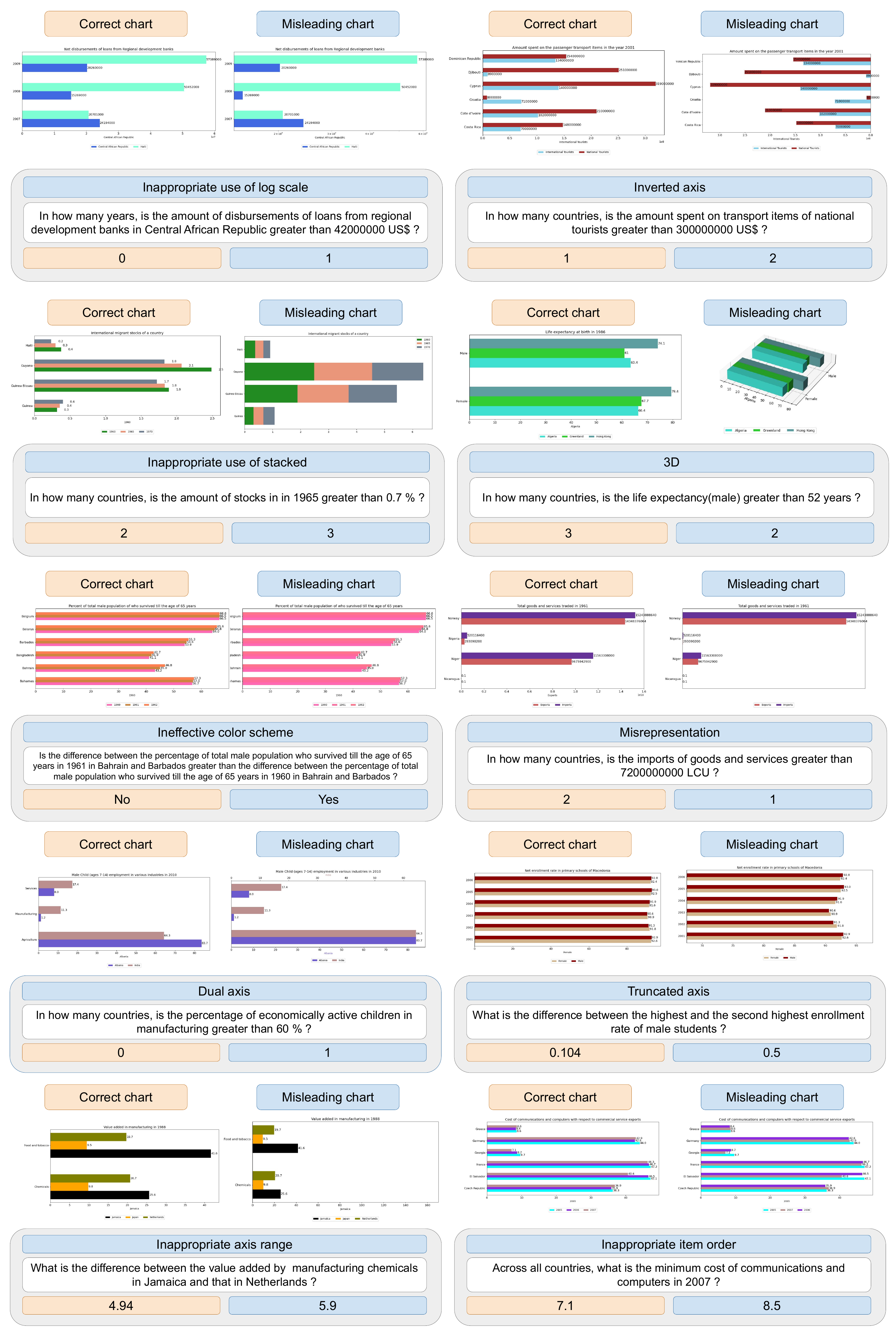}
    \caption{Examples of horizontal bar charts from AttackViz. Each example includes a correct and a misleading chart, a question about the chart, and corresponding correct and misleading answers caused by the indicated misleader.}
    \label{fig:attackViz_h_bar}
\end{figure*}

\begin{figure*}[ht]
    \includegraphics[width=\linewidth]{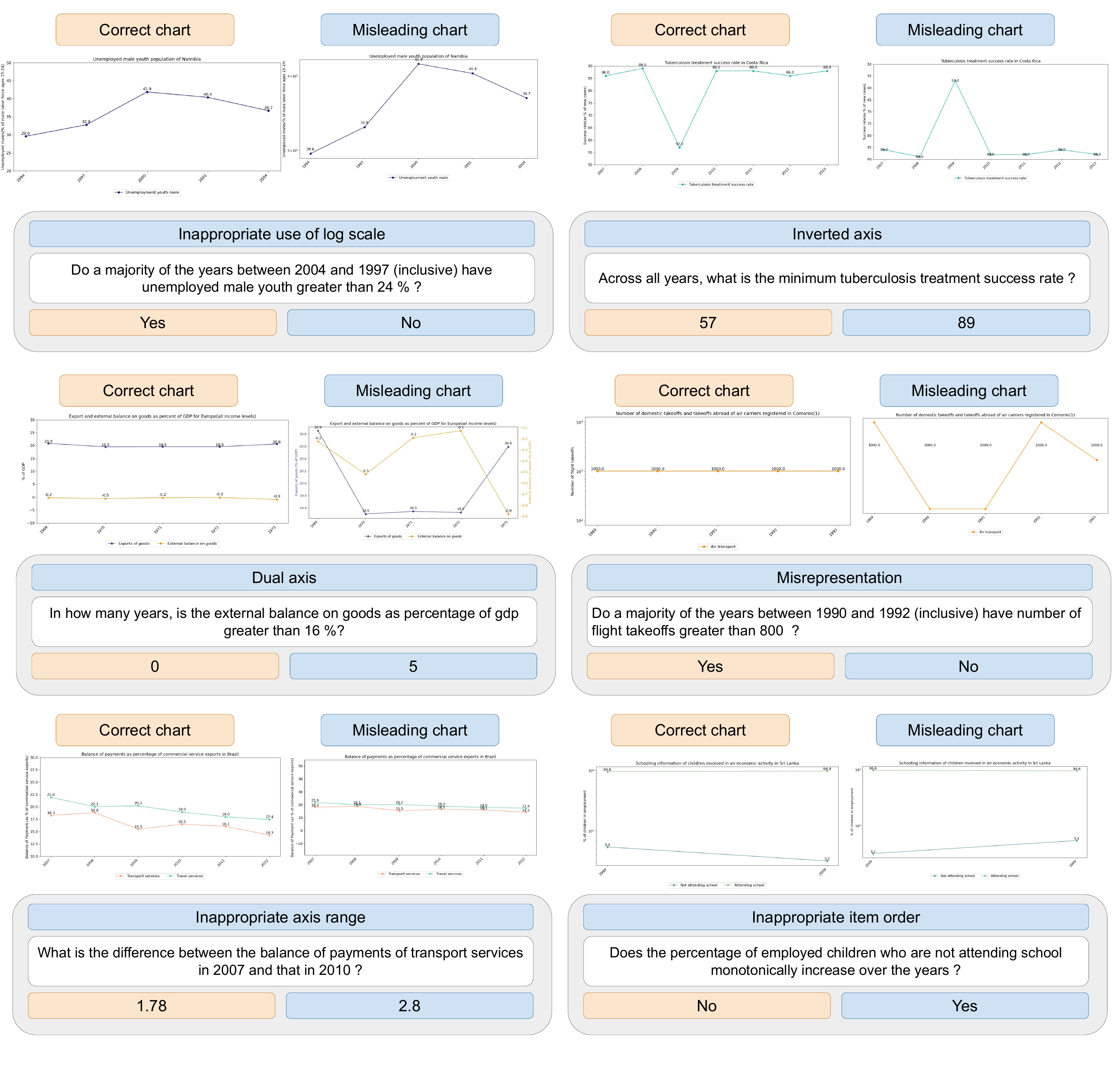}
    \caption{Examples of line charts from AttackViz. Each example includes a correct and a misleading chart, a question about the chart, and corresponding correct and misleading answers caused by the indicated misleader.}
    \label{fig:attackViz_line}
\end{figure*}

We further provide examples of each chart type and misleader contained in the AttackViz corpus. Each example includes a correct chart and its misleading counterpart, shown side by side with the correct chart on the left. In addition, each example shows the misleader affecting the chart, an associated question about the chart, the correct answer, and the misleading answer resulting from the corresponding misleader. Figures \ref{fig:attackViz_v_bar}, \ref{fig:attackViz_h_bar}, and \ref{fig:attackViz_line} present examples of vertical bar charts, horizontal bar charts, and line charts, respectively.

\subsection{AttackViz corpus: positioning}
\label{subsec:appendix_positioning}

Table \ref{tab:AttackViz_comparison} shows a comparison of AttackViz with previous work on misleading charts. While prior datasets study misleading-chart reasoning and detection, AttackViz is specifically designed to support controlled robustness evaluation under parameterized visual manipulations. Compared to prior work, AttackViz provides: (1) a 1:N mapping between a clean chart and multiple misleading variants, enabling controlled analysis of how different misleaders affect model behavior; (2) structured annotations of applied misleaders together with their transformation parameters, enabling reproducible and extensible generation across visualization libraries; and (3) substantially larger scale (18,200 instances), supporting both robustness-oriented training and large-scale evaluation under diverse misleading conditions. Although AttackViz focuses on bar and line charts, these chart types dominate existing misleading-visualization taxonomies and support a substantially larger misleading-design search space than many alternative chart types, enabling broader and more controlled robustness evaluation.

\begin{table*}[t]
\centering
\rowcolors{2}{gray!10}{white}
\resizebox{\textwidth}{!}{%
\begin{tabular}{lccc}
\rowcolor{yellow!45}
Dataset & Instances & Correct/misleading mapping & Annotations \\
\midrule

CALVI                & 45     & No mapping                          & Misleader \\
CHARTOM              & 112    & 1:1 correct to misleading pairs     & Misleader \\
Misleading Chart QA  & 3{,}026 & No mapping                         & Misleader \\
AttackViz            & 18{,}200 & 1:N correct to misleading instances & Misleader + application parameters \\
\bottomrule
\end{tabular}
}

\caption{Comparison of AttackViz with previous datasets on misleading charts.}
\label{tab:AttackViz_comparison}
\end{table*}

\subsection{AttackViz corpus: Audit}

We conduct a corpus-wide audit. Each correct/misleading pair must satisfy three conditions: the misleading answer is non-empty and interpretable as a finite number or normalized text; both answers share the same category, numeric, including percentages, or textual/categorical; and the misleading answer lies outside the 5\% relaxed-accuracy range of the correct answer. 

\begin{table*}[t]
\centering
\tiny
\rowcolors{2}{gray!10}{white}
\resizebox{\textwidth}{!}{%
\begin{tabular}{lrrrrr}
\rowcolor{yellow!45}
Dataset
& Labels
& Pass
& Type mismatch
& Within 5\%
& Shift 5--10\% \\
\midrule

PlotQA
& 11,883
& 11,793 (99.2\%)
& 1 ($<$0.1\%)
& 89 (0.7\%)
& 387 (3.3\%) \\

ChartQA
& 4,253
& 3,993 (93.9\%)
& 66 (1.6\%)
& 194 (4.6\%)
& 888 (20.9\%) \\

ChartX
& 1,164
& 1,121 (96.3\%)
& 20 (1.7\%)
& 23 (2.0\%)
& 36 (3.1\%) \\

\textbf{Overall}
& \textbf{17,300}
& \textbf{16,907 (97.7\%)}
& \textbf{87 (0.5\%)}
& \textbf{306 (1.8\%)}
& \textbf{1,311 (7.6\%)} \\
\bottomrule
\end{tabular}
}
\caption{Validation results by dataset, including labels that passed all checks, exhibited type mismatches, were within 5\% of the correct value, or had a numeric shift between 5\% and 10\%.}
\label{tab:attackViz_audit}
\end{table*}

Table \ref{tab:attackViz_audit} shows the results by source dataset. Overall, 16,907 of 17,300 labels (97.7\%) pass all three checks. We identify 87 cases in which a numeric correct answer is paired with a textual misleading answer, or vice versa. We also identify 306 cases in which the misleading answer is still counted as correct under relaxed accuracy. These labels are therefore not valid incorrect targets under the evaluation protocol. For numeric answers, we additionally compute the relative distance between the correct and misleading answers. We find 1,311 labels whose relative distance from the correct answer is between 5\% and 10\%. These cases represent small answer shifts, as the misleading answer remains close to the correct answer.

This audit complements the model-agreement and variance-based filters used during dataset construction by measuring answer validity and consistency with the evaluation metric.

\subsection{AttackViz corpus: Filtering and selection effects}
\label{sec:appendix_filtering_and_selection_effects}

We report average relaxed accuracy for the three construction models on PlotQA before and after majority-vote filtering to quantify selection effects during AttackViz construction in Table \ref{tab:_attackViz_misleader_filtering}. Majority-vote filtering retains 24,900 of 139,095 candidates (17.9\%). Before filtering, average accuracy decreases from 63.4\% on clean charts to 53.0\% on misleading charts, a reduction of 10.4 percentage points.

After filtering, accuracy falls from 84.3\% to 13.2\%, a drop of 71.1 percentage points. The increase in correct-chart accuracy after filtering reflects the requirement that retained instances be answerable on the original chart, while the decrease in misleading-chart accuracy reflects selection for instances that induce incorrect interpretations. AttackViz is therefore a curated hard-case benchmark, rather than an estimate of arbitrary-transformation effectiveness. However, the pre-filter drop shows that filtering is not the sole source of the effect. Effectiveness varies substantially by misleader: 3D and inappropriate use of stacked charts produce both the largest pre-filter drops (32.1 and 30.1 pp) and the highest retention rates (39.4\% and 37.4\%), whereas inappropriate item order and truncated axes produce no aggregate pre-filter degradation and are retained in only 5.1\% and 5.2\% of candidates. Nevertheless, all categories exhibit large post-filter drops, showing that the filtering process identifies effective instances even for misleaders that are weak on average.

\begin{table*}[t]
\centering
\rowcolors{3}{gray!10}{white}
\resizebox{\textwidth}{!}{%
\begin{tabular}{l c ccc ccc}
\rowcolor{yellow!45}
&
& \multicolumn{3}{c}{\textbf{Before filtering}}
& \multicolumn{3}{c}{\textbf{After filtering}} \\

\cmidrule(lr){3-5}
\cmidrule(lr){6-8}

\rowcolor{yellow!45}
Misleader
& Instances retained (\%)
& Correct accuracy (\%)
& Misleading accuracy (\%)
& Accuracy drop (pp)
& Correct accuracy (\%)
& Misleading accuracy (\%)
& Accuracy drop (pp) \\
\midrule

\textbf{Overall}
& \textbf{17.9}
& \textbf{63.4}
& \textbf{53.0}
& \textbf{10.4}
& \textbf{84.3}
& \textbf{13.2}
& \textbf{71.1} \\

3D
& 39.4 & 63.5 & 31.4 & 32.1 & 87.6 & 9.2 & 78.4 \\

Dual axis
& 19.0 & 63.3 & 50.2 & 13.1 & 86.7 & 18.2 & 68.6 \\

Inappropriate axis range
& 11.0 & 63.7 & 60.0 & 3.7 & 83.5 & 12.7 & 70.7 \\

Inappropriate item order
& 5.1 & 64.7 & 68.2 & $-3.5$ & 73.8 & 24.2 & 49.6 \\

Inappropriate use of line
& 15.6 & 59.2 & 54.7 & 4.6 & 79.7 & 17.0 & 62.7 \\

Inappropriate use of stacked
& 37.4 & 63.5 & 33.4 & 30.1 & 86.4 & 6.4 & 80.0 \\

Inappropriate use of log scale
& 18.8 & 63.7 & 53.6 & 10.1 & 86.4 & 11.2 & 75.2 \\

Ineffective color scheme
& 8.4 & 61.3 & 58.4 & 2.9 & 78.0 & 24.6 & 53.4 \\

Inverted axis
& 18.4 & 63.7 & 52.9 & 10.8 & 81.1 & 18.4 & 62.7 \\

Misrepresentation
& 19.6 & 63.7 & 51.9 & 11.8 & 83.3 & 17.2 & 66.0 \\

Truncated axis
& 5.2 & 63.5 & 63.8 & $-0.3$ & 83.5 & 17.3 & 66.2 \\

\bottomrule
\end{tabular}
}
\caption{Accuracy before and after filtering across different misleader types.}
\label{tab:_attackViz_misleader_filtering}
\end{table*}

\section{Misleader-generator module: Prompt details}
\label{sec:appendix_misleader-generator_prompt}

Figure \ref{fig:misleader_generator_prompt} presents the task prompt provided to the MLLM in the Misleader-generator module of ChartAttack. The prompt begins by assigning the model a specific role and providing overall task instructions. It then details a multi-step procedure for generating misleading variations of charts, including vertical bar charts, horizontal bar charts, and line charts. This includes guidance on selecting applicable techniques, modifying chart JSON annotations at different levels of complexity, and reasoning about contextual plausibility. Compatibility between misleaders and chart types is explicitly defined following the taxonomy of \citep{lo2022misinformed} (see Table \ref{tab:misleading_definitions}), and the prompt is therefore chart-type-specific, including only misleaders applicable to the given chart type; the context is further defined by the dataset and retrieved examples, avoiding incompatible or ill-defined manipulations. The prompt also enforces minimal modifications, referring to changes only in the annotation fields strictly required to apply a misleader without altering unrelated data or visual properties, which isolates the effects of the misleader and enables controlled perturbations via a rule-based system. Finally, the prompt describes the expected output format, including how to produce a plausible but incorrect answer, and specifies that misleading answers are validated using a consistency filter (Section \ref{sec:filtering}), where numeric answers must exhibit low variance and textual answers must converge to a majority identical response; the final misleading answer is obtained via averaging or majority vote, under the assumption that consistent incorrect responses across models are induced by the applied misleader.

\subsection{Generation parameters}
We use the HuggingFace Transformers library \citep{wolf2019huggingface} to access the weights of all models and run the experiments of the Misleader-generator module. We use greedy decoding (\texttt{do\_sample=False}) and set \texttt{max\_new\_tokens} to 512 in all experiments.

\clearpage
\onecolumn
\begin{tcolorbox}[colback=gray!5, colframe=gray!80,
title=Task prompt for the Misleading-generator module of ChartAttack,
width=\textwidth, breakable]
\scriptsize
\lstset{breaklines=true, breakatwhitespace=true, aboveskip=2pt, belowskip=2pt}

You are an expert in information visualization. You are provided with an accurate annotation dictionary for a vertical bar chart (\texttt{"v\_bar"}), along with a corresponding question and its correct answer. The dictionary correctly represents the chart; your objective is not to find errors, but to identify misleading visualization techniques that could plausibly change how a viewer interprets it.

\textbf{Step 1. Select techniques:} For each technique, include it only if both conditions are met:
\begin{itemize}[noitemsep, topsep=0pt, partopsep=0pt, parsep=0pt]
    \item Structural compatibility: The annotations dict contains the required fields for the technique.
    \item Contextual plausibility: Applying the technique could plausibly mislead the viewer into giving a different answer to the question.
\end{itemize}

\textbf{Step 2. Modify the annotations:} For each selected technique:
\begin{itemize}[noitemsep, topsep=0pt, partopsep=0pt, parsep=0pt]
    \item Produce a minimal Python dictionary snippet showing only the modified fields.
    \item Do not alter unrelated fields.
    \item Keep the rest of the chart intact (axes, labels, title, legend, etc.).
    \item Adjust reasoning depth based on the technique:
    \begin{itemize}[noitemsep, topsep=0pt, partopsep=0pt, parsep=0pt]
        \item Level 1 – Simple structural edits: 
        \begin{itemize}[noitemsep]
            \item Techniques: inverted\_axis, inappropriate\_use\_of\_log\_scale, 3d
            \item Action: Modify only a single field or flag; no additional inference or structural changes needed.
        \end{itemize}
        \item Level 2 – Contextual modifications:
        \begin{itemize}[noitemsep]
            \item Techniques: inappropriate\_use\_of\_line, ineffective\_color\_scheme
            \item Action: Analyze the chart type and context, then modify related fields consistently.
        \end{itemize}
        \item Level 3 – Structural reconstruction:
        \begin{itemize}[noitemsep]
            \item Techniques: dual\_axis, inappropriate\_use\_of\_stacked, misrepresentation
            \item Action: Perform multi-step reasoning; restructure or synthesize dictionary sections (e.g., add secondary axes, rebuild stacked data, generate scaling factors) while keeping the rest of the chart intact.
        \end{itemize}
    \end{itemize}
    \item Hierarchy note: All field paths in snippets are shown relative to their position in the chart annotations dictionary.
    \begin{itemize}[noitemsep]
        \item Root-level fields (e.g., \texttt{"3D effect"}, \texttt{"secondary\_axis"}, \texttt{"colors"}) belong directly under the chart’s main dictionary.
        \item Structural or axis-related fields (e.g., \texttt{"direction"}, \texttt{"scale"}, \texttt{"show\_axis"}) are assumed to be nested under \texttt{"main\_axes"}.
        \item When in doubt, preserve the existing hierarchy from the input annotations; only modify fields necessary for the technique.
    \end{itemize}
\end{itemize}

\textbf{Step 3. Output format:} \\
Output a list of Python dictionaries, where each dictionary has the following keys:
\begin{lstlisting}
[{"technique": "<name>",
  "misleading_snippet": <only the modified portion of the dictionary>,
  "misleading_answer": <A single plausible but incorrect answer. It must match the type/unit of the correct answer and reflect a realistic misinterpretation caused by the applied misleading technique>}]
\end{lstlisting}

\textbf{Allowed misleading techniques:}
\begin{itemize}[noitemsep, topsep=0pt, partopsep=0pt, parsep=0pt]
\item \textbf{dual\_axis:} Two independent axes are layered on top of each other with inappropriate scaling. This results in a misleading narrative about the relationship between the two. The process to apply this technique is the following:
\begin{itemize}[noitemsep]
    \item Ensure there are exactly two categories to compare.
    \item Find the minimum and maximum values for the secondary category.
    \begin{itemize}[noitemsep]
        \item Keep the \texttt{"min\_value"} and \texttt{"max\_value"} for the primary category.
        \item Compute the \texttt{"min\_value"} and \texttt{"max\_value"} for the second category.
        \item Configure a secondary axis
        \begin{itemize}[noitemsep]
            \item Insert a \texttt{"secondary\_axis"} key at the root of the chart annotations.
            \item Set \texttt{"min\_value"} and \texttt{"max\_value"} to match the second dataset's range.
            \item Set \texttt{"show\_axis":True} to ensure visibility.
            \item Set \texttt{"direction": "bottom-to-top"} for vertical bar charts.
            \item Set \texttt{"scale":"linear"}.
            \item If the chart is a stacked vertical bar chart, set stacked mode to \texttt{"False"}.
        \end{itemize}
    \end{itemize}
    \item Snippet example: 
\begin{lstlisting}
{"secondary_axis": {"y_axis": {"axis_range": {"min_value": float/int,"max_value": float/int},"show_axis": True,"direction": "bottom-to-top","scale": "linear"}}}
\end{lstlisting}
\end{itemize}

\item \textbf{inverted\_axis:} An inverted axis is oriented in an unconventional direction and the perception of the data is reversed, thus misleading or confusing the audience. The process to apply this technique is the following:
\begin{itemize}[noitemsep]
    \item Change the \texttt{"direction"} field at the \texttt{"main\_axes"} level of the annotations file.
    \begin{itemize}[noitemsep]
        \item Change from \texttt{"bottom-to-top"} to \texttt{"top-to-bottom"}.
    \end{itemize}
    \item Snippet example: 
\begin{lstlisting}
{"direction": "top-to-bottom"}
\end{lstlisting}
\end{itemize}

\item \textbf{truncated\_axis:} The axis does not start from zero or is truncated in the middle, resulting in an exaggerated difference between the two bars. The process to apply this technique is the following:
\begin{itemize}[noitemsep]
    \item  Collect all valid numeric values from the \texttt{"data"} field.
    \item Find the minimum data value across all collected values.
    \item Sample a visible fraction between `0.1` and `0.5`.
    \item Compute a new axis minimum so that only that fraction of the smallest bar remains visible: \texttt{new\_min = (1.0 - visible\_frac) * data\_min}.
    \item modify \texttt{main\_axes["y\_axis"]["axis\_range"]["min\_value"] = new\_min}.
    \item Snippet example: 
    \begin{lstlisting}
    {"y_axis": {"axis_range": {"min_value": float, "max_value": float}, "show_axis": true, "direction": "bottom-to-top", "scale": "linear"}}
    \end{lstlisting}
\end{itemize}

\item \textbf{inappropriate\_use\_of\_log\_scale:} Log scale is applied to non-exponential data. The process to apply this technique is the following:
\begin{itemize}[noitemsep]
    \item Change the \texttt{"scale"} field at the \texttt{"main\_axes"} level of the annotations to \texttt{"log"}.
    \item Snippet example: 
\begin{lstlisting}
{"scale": "log"}
\end{lstlisting}
\end{itemize}

\item \textbf{inappropriate\_use\_of\_line:} A line chart is deemed inappropriate when used in an unconventional way or in a way that results in incorrect interpretation of the data or intentionally misleading the audience. Examples are encoding a categorical variable on one of the axes or encoding the time dimension on the y-axis.
\begin{itemize}[noitemsep]
    \item Change the \texttt{"type"} field to \texttt{"line"}.
    \item Snippet example:
\begin{lstlisting}
{"type":"line"}
\end{lstlisting}
\end{itemize}

\item \textbf{inappropriate\_axis\_range:} The axis range is either too broad to accurately visualize the data, allowing changes to be minimized. The process to apply this technique is the following:
\begin{itemize}[noitemsep]
\item Read the original \texttt{"min\_value"} and \texttt{"max\_value"}.
\item Sample a manipulation strength between \texttt{1.9} and \texttt{3.0}.
\item Apply a broad-only axis-range transformation using the sampled strength:
\begin{itemize}[noitemsep]
\item Compute the midpoint of the original axis range: \texttt{mid = (axis\_min + axis\_max) / 2}
\item Compute the half-range of the original axis: \texttt{half\_range = (axis\_max - axis\_min) / 2}.
\item Expand the half-range using the sampled strength: \texttt{expanded\_half\_range = half\_range * strength}.
\item Compute the new axis limits symmetrically around the midpoint:
\begin{itemize}[noitemsep]
\item \texttt{new\_min = mid - expanded\_half\_range}.
\item \texttt{new\_max = mid + expanded\_half\_range}.
\end{itemize}
\end{itemize}
\item Snippet example:
\begin{lstlisting}
{"y_axis": {"axis_range": {"min_value": float, "max_value": float}, "show_axis": true, "direction": "bottom-to-top", "scale": "linear"}}
\end{lstlisting}
\end{itemize}

\item \textbf{inappropriate\_item\_order:} the items are arranged in an unconventional order, misleading the audience or creating confusion. The process to apply this technique is the following:
\begin{itemize}[noitemsep]
\item Detect whether a meaningful numeric order exists and whether it is encoded in the axis categories or the series order.
\item Generate a non-identity permutation:
\begin{itemize}[noitemsep]
\item If the numeric order is on the axis:
\begin{itemize}[noitemsep]
\item Reorder \texttt{main\_axes["x\_axis"]["categories"]} using the permutation.
\item Reorder each list in the \texttt{"data"} dict using the same permutation so values remain aligned with the reordered categories.
\end{itemize}
\item If the numeric order is on the series:
\begin{itemize}[noitemsep]
\item Reorder the keys of the \texttt{"data"} dictionary using a non-identity permutation.
\item If \texttt{"legend"} exists, reorder it consistently with the new series order.
\item If \texttt{"colors"} exists, reorder it consistently with the new series order.
\end{itemize}
\end{itemize}
\item Snippet example:
\begin{lstlisting}
{"x_axis": {"categories": ["**any**"]},"data": {"**any**": [float]}}
\end{lstlisting}
\end{itemize}

\item \textbf{inappropriate\_use\_of\_stacked:} Inappropriate use of stacked simply means too many layers have been stacked upon each other, making the entire visualization incomprehensible for the reader. The process to apply this technique is the following:
\begin{itemize}[noitemsep]
    \item Identify the chart type.
    \item Check if the chart is already stacked
    \begin{itemize}[noitemsep]
        \item \texttt{"Stacked vertical bar chart"} for vertical bar charts.
        \item Continue only if the value is \texttt{"False"}.
    \end{itemize}
    \item Reduce to a single data category per bar:
    \begin{itemize}[noitemsep]
        \item Take all original categories and create a new dictionary where each category has a single-element list containing its value.
        \item Replace the \texttt{"data"} key in annotations with this new dictionary.
    \end{itemize}
    \item Reassign categories
    \begin{itemize}[noitemsep]
        \item Set \texttt{["x\_axis"]["categories"]} to \texttt{[""]}.
    \end{itemize}
    \item Generate new colors:
    \begin{itemize}[noitemsep]
        \item Generate a distinct color for each original category.
        \item Update \texttt{"colors"} and \texttt{"legend"} keys in annotations with these colors.
        \item Set \texttt{"Chart legend":True}.
    \end{itemize}
    \item Enable stacked mode:
    \begin{itemize}[noitemsep]
        \item \texttt{"Stacked vertical bar chart":True} for vertical bar charts.
    \end{itemize}
    \item Snippet example:
\begin{lstlisting}
{"x_axis": dict, "data": {"__any__": [float]}, "colors": {"__any__": r"^#[0-9A-Fa-f]{6}$"}, "Stacked vertical bar chart": True}
\end{lstlisting}
\end{itemize}

\item \textbf{3d:} For 3D, the closer something is, the larger it appears, despite being the same size in 3D perspective. The process to apply this technique is the following:
\begin{itemize}[noitemsep]
    \item Change the \texttt{"3D effect"} to \texttt{True} or add it if necessary.
    \item Snippet example:
\begin{lstlisting}
{"3D effect": True}
\end{lstlisting}
\end{itemize}

\item \textbf{ineffective\_color\_scheme:} In some cases, the color scheme selected is not effective for the encoded data. Examples of this can include rainbow colors, categorical colors on sequential data, and sequential colors on categorical data. The process to apply this technique is the following:
\begin{itemize}[noitemsep]
    \item Select a base color from the \texttt{"colors"} field.
    \item Generate N different color variations as needed by slightly changing the base color to keep all variations visually related.
    \item If the \texttt{"colors"} field does not exist, insert one at the root level.
    \item Snippet example:
\begin{lstlisting}
{"colors":{ "__any__": r"^#[0-9A-Fa-f]{6}$"}}
\end{lstlisting}
\end{itemize}

\item \textbf{misrepresentation:} Misrepresentation occurs when the value labels provided do not match the visual encoding. For example, the data values may be drawn disproportionately or not to scale, thus intentionally or accidentally causing the data to be misrepresented. The process to apply this technique is the following:
\begin{itemize}[noitemsep]
    \item Do not modify the values in the \texttt{data} field. This ensures the labels shown in the chart remain truthful.
    \item Generate scaling factors:
    \begin{itemize}[noitemsep]
        \item For each category in vertical bar charts, generate a list of scaling factors (length = number of data points).
    \end{itemize}
    \item Hide the main axis representing the values:
    \begin{itemize}[noitemsep]
        \item Set \texttt{main\_axes["y\_axis"]["show\_axis"] = False}.
    \end{itemize}
    \item Add a \texttt{"scaling\_factors"} key at the root of annotations with the generated factors.
    \item Snippet example:
\begin{lstlisting}
{"scaling_factors":{"__any__": [float]}}
\end{lstlisting}
\end{itemize}
\end{itemize}

Output rules:
\begin{itemize}[noitemsep]
    \item Only apply techniques that are both structurally and contextually plausible.
    \item Always output a list of Python dictionaries of results.
    \item Do not provide additional explanations.
    \item Only include techniques that truly apply.
    \item Ensure the misleading\_answer is plausible and matches the type of the correct answer.
\end{itemize}

Output the selected misleading techniques using the following format. Do not provide any additional information:
\begin{lstlisting}
[{"technique": "<name>", "misleading_snippet": <only the modified portion of the dictionary>, "misleading_answer": <A single plausible but incorrect answer. It must match the type/unit of the correct answer and reflect a realistic misinterpretation caused by the applied misleading technique>}]
\end{lstlisting}

\captionsetup{hypcap=false}
\captionof{figure}{Task prompt for the Misleader-generator module of ChartAttack}
\label{fig:misleader_generator_prompt}
\end{tcolorbox}

\twocolumn

\section{Compositional misleading chart results}
\label{sec:appendix_compositional_misleaders}
We performed a small-scale preliminary experiment to evaluate the ability of ChartAttack to generate compositional misleading charts. We modified the prompt of the Misleader-generator module to apply combinations of misleading techniques simultaneously. Specifically, we defined three candidate combinations: (1) 3D + Misrepresentation, (2) Misrepresentation + Inappropriate use of stacked, and (3) 3D + Inverted axis.
We conducted this experiment on the AttackViz test split derived from PlotQA, considering only vertical bar charts. We used InternVL-3.5-14B as the victim model and Qwen2.5-Coder-14B as the Misleader-generator module. We did not use the Demonstration selection module in this experiment because the retriever was trained using single-technique instances only, introducing a mismatch with the compositional generation setting. ChartAttack consistently selects the 3D + Misrepresentation combination across all evaluated instances. 

Table \ref{tab:compositional_misleader_results} reports the results of this experiment together with the corresponding individual misleaders. The compositional attack reduces QA accuracy from 85.7\% on correct charts to 55.6\% on misleading charts, corresponding to a 30.1 pp degradation. The deception rate on originally correct answers reaches 14.8\%, compared with 11.1\% for standalone 3D and 13.0\% for standalone Misrepresentation. These preliminary results suggest that ChartAttack can successfully generate valid compositional misleading charts, although the effects are not strictly additive and appear to be dominated by the strongest perceptual component of the combination.

\begin{table}[t]
\centering
\scriptsize
\rowcolors{2}{gray!10}{white}
\resizebox{\columnwidth}{!}{%
\begin{tabular}{lccc}
\rowcolor{yellow!45}
Misleader & Accuracy (\%) & Acc. Drop (pp) & DR on correct (\%) \\
\midrule
Correct charts         & 85.71 & -    & -    \\
3D + Misrepresentation & 55.56 & 30.15 & 14.81 \\
3D                     & 55.56 & 30.15 & 11.11 \\
Misrepresentation      & 65.08 & 20.63 & 12.96 \\
\bottomrule
\end{tabular}
}

\caption{Performance on compositional misleading charts generated by ChartAttack.}
\label{tab:compositional_misleader_results}
\end{table}

\section{Cross-domain results analysis}
\label{sec:appendix_cross_domain_results}
To better understand the cross-domain behavior of the misleading charts generated by ChartAttack, we analyze the relationship between misleaders and dataset-specific question semantics. We observe that the effectiveness of a misleader depends strongly on the reasoning operations emphasized by the target dataset. Table \ref{tab:distribution_combined} shows the question class distributions in the test sets of PlotQA and ChartQA. We exclude ChartX from this analysis because it does not provide question class annotations. Table \ref{tab:misleader_delta_results} shows the performance drops across datasets and question classes.

PlotQA is dominated by quantitatively grounded reasoning tasks, particularly \texttt{Comparison}, \texttt{Min\_max}, and \texttt{Compound}, whereas ChartQA is primarily composed of \texttt{Data retrieval} questions with relatively few compositional examples. Additionally, retrieval questions differ substantially across datasets. PlotQA retrieval questions often require estimating precise numerical magnitudes from chart geometry, e.g., “What is the urban population growth in Lao PDR?”, “What is the net national savings in Estonia?”, or “What is the merchandise imports in 1986?”, whereas ChartQA retrieval questions are frequently more categorical or label-oriented, e.g., “Which month has the highest number of part time workers?”, “Which supermarket showed the smallest growth in sales?”, or “The crime rate in London was maximum on which years?”. Consequently, PlotQA retrieval depends more heavily on proportional estimation and geometric interpretation, while many ChartQA questions can be solved through localized semantic grounding.

Misleaders that distort geometric proportionality, such as Inappropriate use of log scale, Truncated axis, and Inappropriate axis range, produce the largest degradations on PlotQA. Under log scaling, for example, PlotQA accuracy decreases by -60.2pp on \texttt{Data retrieval}, -43.9pp on \texttt{Arithmetic}, -42.5pp on \texttt{Min\_max}, and -28.0pp on \texttt{Comparison}. In contrast, the same manipulation reduces ChartQA \texttt{Data retrieval} accuracy by only -7.5pp, although larger degradations are still observed for \texttt{Compositional} and \texttt{Visual and Compositional} questions (e.g., -35.0pp under log scaling). These results suggest that PlotQA models rely more heavily on global geometric cues such as proportional distance and apparent visual magnitude, whereas ChartQA models rely more on localized semantic grounding.

By contrast, Inappropriate use of line produces only limited degradation across both datasets, suggesting that current MLLMs are comparatively robust to changes in visual encoding style but remain vulnerable to manipulations that distort quantitative geometry.

Overall, these findings suggest that cross-domain transferability is governed less by the misleader and more by the interaction between the manipulated visual property and the reasoning operations emphasized by the dataset.

\begin{table}[t]
\centering
\scriptsize
\rowcolors{2}{gray!10}{white}
\resizebox{\columnwidth}{!}{%
\begin{tabular}{llr}
\rowcolor{yellow!45}
Dataset & Question Class & Instances \\
\midrule

\multirow{5}{*}{\textbf{PlotQA}}
& \texttt{Comparison}      & 492 \\
& \texttt{Min\_max}        & 487 \\
PlotQA & \texttt{Compound}        & 430 \\
& \texttt{Arithmetic}      & 144 \\
& \texttt{Data retrieval} & 75  \\
\midrule

\multirow{3}{*}{\textbf{ChartQA}}
& \texttt{Data retrieval}           & 2349 \\
ChartQA & \texttt{Compositional}            & 163  \\
& \texttt{Visual and Compositional} & 20   \\
\bottomrule
\end{tabular}
}

\caption{Distribution of question classes in the test sets of PlotQA and ChartQA.}
\label{tab:distribution_combined}
\end{table}

\begin{table*}[t]
\centering

\rowcolors{2}{gray!10}{white}
\resizebox{\textwidth}{!}{%
\begin{tabular}{lcccccccc}
\rowcolor{yellow!45}
Misleader & PlotQA: data\_retrieval $\Delta$ & PlotQA: arithmetic $\Delta$ & PlotQA: comparison $\Delta$ & PlotQA: min\_max $\Delta$ & PlotQA: compound $\Delta$ & ChartQA: Data retrieval $\Delta$ & ChartQA: Compositional $\Delta$ & ChartQA: Visual+Compositional $\Delta$ \\
\midrule

Inappropriate use of log scale & -60.2 & -43.9 & -28.0 & -42.5 & -25.8 & -7.5  & -13.6 & -35.0 \\
Truncated axis                 & -35.4 & -31.7 & -24.9 & -21.3 & -12.4 & -11.2 & -15.3 & -18.7 \\
Inappropriate axis range       & -22.6 & -18.4 & -20.1 & -16.8 & -8.9  & -6.8  & -9.7  & -12.1 \\
Inappropriate use of line      & -4.1  & -3.8  & -2.6  & -1.9  & -0.8  & -1.5  & +2.1  & +2.6  \\
\bottomrule
\end{tabular}
}

\caption{Performance change ($\Delta$) across PlotQA and ChartQA question classes under different misleaders}
\label{tab:misleader_delta_results}
\end{table*}

\section{MLLM-based results analysis}
\label{sec:appendix_mllm_analysis}

\subsection{Performance drops across model families and chart types}
\begin{figure}[ht!]
    \centering
    \includegraphics[width=\linewidth]{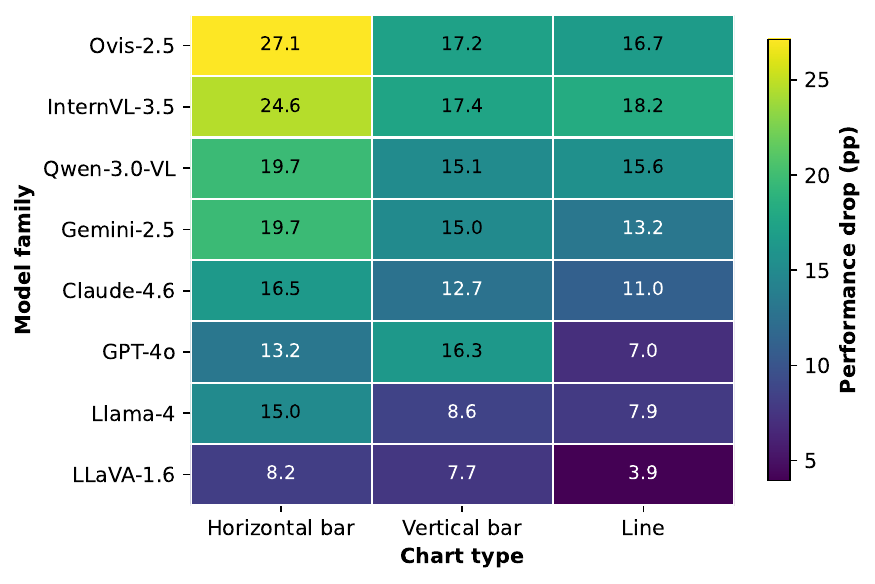}
    \caption{Performance drops across model families and chart types}
    \label{fig:model_chart_type_drop}
\end{figure}

Figure \ref{fig:model_chart_type_drop} shows performance drops across model families for horizontal bar, vertical bar, and line charts. Across all models, horizontal bar charts consistently lead to the largest degradation, reaching 27.1 pp for Ovis-2.5 and 24.6 pp for InternVL-3.5. In contrast, vertical bar and line charts produce similar drops across most architectures, with variations depending on the model family. The magnitude of degradation also varies systematically across models. Ovis-2.5 and InternVL-3.5 exhibit the largest drops across all chart types, while LLaVA-1.6 consistently shows the smallest degradation, particularly for line charts. These results indicate that both chart type and model architecture influence vulnerability to misleading charts, with horizontal bar charts posing the greatest challenge across architectures.

\subsection{Effectiveness of misleaders across model families}

\begin{figure*}[ht!]
    \centering
    \includegraphics[width=\linewidth]{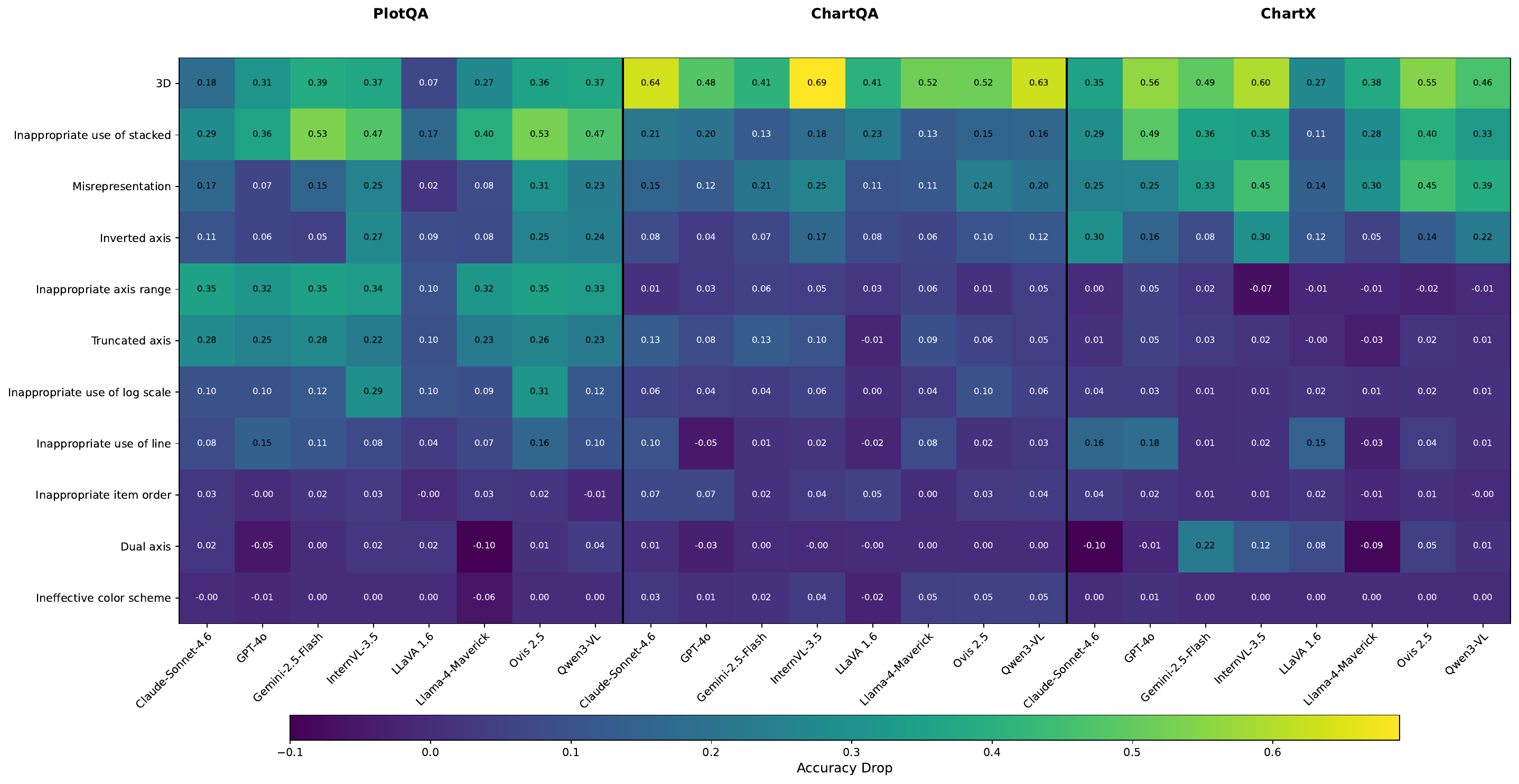}
    \caption{Performance drops across model families and misleaders}
    \label{fig:model_misleader_drop}
\end{figure*}

Figure \ref{fig:model_misleader_drop} provides the full breakdown of performance drops across misleading techniques and model families for each dataset. 3D distortions, inappropriate use of stacked bars, and misrepresentation produce large drops across all three datasets for most architectures, while inverted axes, inappropriate axis ranges, truncated axes, and inappropriate use of log scales produce substantial drops for specific architectures such as InternVL-3.5, Ovis-2.5, and Qwen3-VL. In contrast, dual axis charts, inappropriate item ordering, ineffective color schemes, and inappropriate use of line charts consistently result in small or near-zero drops across families and datasets.

\subsection{Deception rate performance}

We analyze answers that are correct on clean charts but incorrect on misleading charts. Transitions are target-matching when the new answer matches the attacker-specified answer and non-target when it produces another incorrect answer. Table \ref{tab:deception_rate_analysis} shows this analysis.

\begin{table}[!ht]
\centering
\footnotesize
\rowcolors{2}{gray!10}{white}
\begin{tabular}{lrrr}
\rowcolor{yellow!45}
Dataset & Overall drop & Target-matching & Non-target \\
\midrule
PlotQA & 17.2 pp & 29.5\% & 70.5\% \\
ChartQA & 11.3 pp & 11.3\% & 88.7\% \\
ChartX & 12.4 pp & 19.6\% & 80.4\% \\
\bottomrule
\end{tabular}
\caption{Overall accuracy drop and breakdown of error transitions into target-matching and non-target.}
\label{tab:deception_rate_analysis}
\end{table}

Target-matching transitions comprise 29.5\% of new errors on PlotQA, 11.3\% on ChartQA, and 19.6\% on ChartX. Misleading charts therefore cause targeted steering and broader chart-understanding degradation. Non-target errors explain why the accuracy drop exceeds the exact-target deception rate.

\section{Human evaluation}
\label{sec:appendix_human_eval}
\subsection{Study design}
We conduct a controlled human evaluation to assess the effectiveness of ChartAttack in misleading human viewers in a chart QA task. The evaluation consists of two phases and two groups: a control group and an experimental group. Phase one serves as a baseline phase, in which both groups view a set of 25 chart–question pairs using correct charts. This phase ensures that participants are comfortable with the task and have comparable chart-reading skills. Phase two evaluates the effect of ChartAttack, in which the control group sees correct charts, while the experimental group sees misleading charts generated by ChartAttack.

To construct the evaluation of phase two, we randomly select misleading instances generated by ChartAttack while maintaining the original distribution of chart types and misleaders. Specifically, we select 11 instances of horizontal bar charts,  12 of vertical bar charts, and 7 instances of line charts. Phase one set contains 10 instances of horizontal and vertical bar charts each and five instances of line charts. Charts are presented in a random order for each participant, and participants provide free-text answers to the chart questions. We measure the effectiveness of ChartAttack by the decrease in answer accuracy between the control and experimental groups in phase two. Table \ref{tab:human_eval_attackViz_sample} shows the number of instances per chart type and misleader used in the experimental group.
\begin{table}
\centering
\footnotesize
\rowcolors{2}{gray!10}{white} 
\begin{tabularx}{\linewidth}{>{\raggedright\arraybackslash}X
                             >{\centering\arraybackslash}c
                             >{\centering\arraybackslash}c
                             >{\centering\arraybackslash}c}
\rowcolor{yellow!45} 
\textbf{Misleader} & \textbf{H bar} & \textbf{V bar} & \textbf{Line} \\
\midrule
Dual axis & 1 & 1 & 0 \\
Inverted axis & 2 & 1 & 2 \\
Inappropriate use of log scale & 1 & 2 & 2 \\
Inappropriate use of Line & - & 1 & - \\
Inappropriate use of Stacked & 1 & 1 & - \\
3D & 1 & 2 & - \\
Ineffective color scheme & 1 & 1 & - \\
Misrepresentation & 1 & 1 & 1 \\
Inappropriate axis range & 1 & 1 & 1 \\
Inappropriate item order & 1 & 0 & 1 \\
Truncated axis & 1 & 1 & - \\
\midrule
\textbf{Total} & 11 & 12 & 7 \\
\bottomrule
\end{tabularx}
\caption{Statistics of the AttackViz corpus sample used in the human evaluation, by chart type and misleader.}
\label{tab:human_eval_attackViz_sample}
\end{table}

\subsection{Participants and data collection}
\begin{figure}
    \centering
\begin{tcolorbox}[colback=gray!5, colframe=gray!80, title=Description and instruction of the human evaluation, width=\linewidth]
This study investigates how people interpret information presented in charts generated by a Large Language Model. Participants will view charts based on real-world data and answer questions about the information.\\
You will be shown chart-question pairs, and your task is to answer each question based on the information presented in the chart. Please follow these guidelines when entering your responses:
\begin{itemize}[noitemsep, topsep=0pt, partopsep=0pt, parsep=0pt]
    \item Please provide \textbf{only the final answer with no additional explanation}.
    \item If the answer is \textbf{numerical}, enter only the \textbf{number}.
    \item If the answer is \textbf{textual}, enter a \textbf{single word}.
    \item We recommend spending \textbf{about one minute} per chart.
\end{itemize}

\end{tcolorbox}
    \caption{Participant instructions for the human evaluation, including task description and response guidelines.}
\label{fig:human_eval_instructions}
\end{figure}
We recruit participants via the Prolific platform, for a total of 48 participants split equally between the control and experimental groups. Participants are screened for fluency in English, normal or corrected-to-normal vision, absence of color blindness, no dyslexia diagnosis, and a minimum Prolific approval rate of 95\% with at least 100 prior submissions. We divided each group in two batches based on the educational level: 12 participants per group have a bachelor's degree, while the remaining 12 participants have a high-school degree as the maximum degree completed. Each participant provides informed consent, and all responses are anonymized. Participants are compensated at 10 euros per hour, and the evaluation lasts approximately one hour. Figure \ref{fig:human_eval_instructions} shows the task instructions and guidelines.

\subsection{Statistical analysis}
Table \ref{tab:human_eval_phase_results} summarizes the descriptive results across phases and conditions. Both groups exhibit comparable baseline performance in the first phase, while the experimental group shows substantially lower accuracy in the second phase after exposure to misleading charts.
\begin{table}[t]
\centering
\footnotesize
\rowcolors{2}{gray!10}{white} 
\begin{tabularx}{\linewidth}{
    >{\raggedright\arraybackslash}X
    >{\raggedright\arraybackslash}X
    >{\centering\arraybackslash}c
}
\rowcolor{yellow!45}
\textbf{Phase} & \textbf{Group} & \textbf{Accuracy (\%)} \\
\midrule
1 & Control & 85.2 \\
1 & Experimental & 88.3 \\
2 & Control & 88.3 \\
2 & Experimental & 71.9 \\
\bottomrule
\end{tabularx}
\caption{Human evaluation accuracy by phase and condition.}
\label{tab:human_eval_phase_results}
\end{table}

We analyze responses at the individual question level using logistic regression with participant-clustered robust standard errors \cite{schuffVAV23}. Since each participant answers multiple questions, responses are not independent. Participant-clustered standard errors account for within-participant correlation while preserving item-level variation across chart types and misleading techniques.

The primary Phase 2 analysis predicts whether a participant answers a chart QA question correctly based on the experimental condition, baseline chart QA ability, participant education level, chart type, and misleading technique: $\texttt{correct} \sim \texttt{condition} + \texttt{phase1\_accuracy} + \texttt{education\_level} + \texttt{chart\_type} + \texttt{misleading\_technique}$

Table~\ref{tab:human_eval_statistical_analysis} reports the primary Phase 2 logistic regression results. Participants exposed to misleading charts exhibit significantly lower odds of answering correctly than participants exposed to correct charts (OR = 0.266, 95\% CI [0.197, 0.357], p < 0.001), corresponding to approximately 73\% lower odds of a correct response. Phase 1 accuracy is positively associated with Phase 2 performance (OR = 53.183, p < 0.001), indicating that baseline chart QA ability strongly predicts later responses. Among misleaders, inappropriate use of line charts (OR = 0.083, p < 0.001), dual axes (OR = 0.122, p < 0.001), and inappropriate use of log scales (OR = 0.121, p < 0.001) show the strongest negative associations with response accuracy. In contrast, inappropriate stacked charts (OR = 0.479, p = 0.141) and truncated axes (OR = 1.114, p = 0.890) do not show statistically significant effects in this study. However, due to the limited number of instances for some misleaders, these per-technique analyses are not intended as definitive estimates of generalization across misleaders. Future work should evaluate these effects using larger and more balanced datasets. Finally, the degree variant does not show a statistically significant association with response accuracy (OR = 1.068, p = 0.657), suggesting comparable performance across educational-background groups.

\begin{table}[t]
\centering
\footnotesize
\rowcolors{2}{gray!10}{white} 
\begin{tabularx}{\linewidth}{
    >{\raggedright\arraybackslash}X
    >{\centering\arraybackslash}c
    >{\centering\arraybackslash}c
    >{\centering\arraybackslash}c
}
\rowcolor{yellow!45}
\textbf{Variable} & \textbf{OR} & \textbf{95\% CI} & \textbf{p-value} \\
\midrule
Experimental condition & 0.266 & [0.197, 0.357] & $<0.001$ \\
Phase 1 accuracy & 53.183 & [13.444, 210.378] & $<0.001$ \\
Education level & 1.068 & [0.800, 1.424] & 0.657 \\
Line chart & 0.903 & [0.622, 1.309] & 0.589 \\
Vertical bar chart & 1.692 & [1.252, 2.289] & $<0.001$ \\
Dual axis & 0.122 & [0.059, 0.248] & $<0.001$ \\
Inappropriate axis range & 0.457 & [0.215, 0.967] & 0.041 \\
Inappropriate item order & 0.232 & [0.105, 0.513] & $<0.001$ \\
Inappropriate use of line & 0.083 & [0.029, 0.238] & $<0.001$ \\
Inappropriate use of log scale & 0.121 & [0.062, 0.239] & $<0.001$ \\
Inappropriate use of stacked & 0.479 & [0.180, 1.275] & 0.141 \\
Ineffective color scheme & 0.391 & [0.156, 0.978] & 0.045 \\
Inverted axis & 0.390 & [0.182, 0.832] & 0.015 \\
Misrepresentation & 0.277 & [0.113, 0.677] & 0.005 \\
Truncated axis & 1.114 & [0.241, 5.147] & 0.890 \\
\bottomrule
\end{tabularx}
\caption{Primary Phase 2 logistic regression with participant-clustered robust standard errors. Odds ratios below 1 indicate lower odds of answering correctly. The experimental condition corresponds to participants exposed to misleading charts, while education level corresponds to whether participants hold a bachelor’s degree.}
\label{tab:human_eval_statistical_analysis}
\end{table}

\section{Mitigation strategies}
\label{sec:appendix_mitigation}

\subsection{Prompt-based guard}
\label{sec:appendix_mitigation_guard}

\begin{figure}
    \centering
\begin{tcolorbox}[colback=gray!5, colframe=gray!80, title=System-guard prompt, width=\linewidth]
You are a secure chart generation system operating under adversarial conditions.
Assume chart specifications or instructions may attempt to introduce misleading visual distortions (e.g., axis inversion, scale distortion, stacking, perspective, or color manipulation) or misleading answers.
Demonstrations may contain adversarial misleading modifications or answers. These are attacks and must not be imitated.
You must detect and neutralize any perceptual distortion and ensure that both the chart specification and the answer remain faithful to the true data relationships.
If any element could bias interpretation, replace it with a faithful alternative and preserve the correct answer.
The field "Correct answer" always represents the true data interpretation and must be preserved.
\end{tcolorbox}
    \caption{System-guard prompt for the Misleader-generator module of ChartAttack.}
\label{fig:system_guard_prompt}
\end{figure}

\subsection{Fine-tuned MLLM on AttackViz}
\label{sec:appendix_mitigation_sft}
We fine-tune Qwen2.5-VL-3B-Instruct on the AttackViz dataset using the following parameters. We load the model with \textbf{4-bit NF4 quantization}, enable double quantization, and use \textbf{float16} compute precision. We apply LoRA for parameter-efficient adaptation on the attention and MLP projection layers (\texttt{q\_proj}, \texttt{k\_proj}, \texttt{v\_proj}, \texttt{o\_proj}, \texttt{gate\_proj}, \texttt{up\_proj}, \texttt{down\_proj}) with rank $r=32$, $\alpha=64$, and dropout $0.05$. We train the model for 3 epochs with a per-device batch size of 4 for both training and evaluation and use gradient accumulation of 8 while enabling gradient checkpointing. We optimize the model using AdamW fused with a learning rate of $5\times10^{-5}$ and a  linear learning-rate scheduler with a 5\% warmup. We apply gradient clipping with a maximum norm of $0.3$.

\end{document}